\documentclass[letterpaper,journal]{IEEEtran}
\usepackage{amsmath,amsfonts}
\usepackage{amssymb}
\usepackage{amsthm}
\usepackage{algorithmic}
\usepackage[linesnumbered,ruled,vlined]{algorithm2e}
\usepackage{array}
\usepackage[caption=false,font=footnotesize,labelfont=rm,textfont=rm]{subfig}
\usepackage{textcomp}
\usepackage{stfloats}
\usepackage{url}
\usepackage{verbatim}
\usepackage{graphicx}
\usepackage{float}
\graphicspath{{./}{figures/}}
\DeclareGraphicsExtensions{.pdf,.jpeg,.png,.eps}

\def\BibTeX{{\rm B\kern-.05em{\sc i\kern-.025em b}\kern-.08em
    T\kern-.1667em\lower.7ex\hbox{E}\kern-.125emX}}
\usepackage{balance}
\usepackage{tabularx}
\usepackage{multirow}

\usepackage{booktabs}
\usepackage{adjustbox}
\usepackage{xcolor}
\usepackage{cite}

\usepackage{color}

\usepackage[pdfpagelabels]{hyperref}
\hypersetup{%
    bookmarksnumbered=true,
    bookmarksopen=true,
    bookmarksopenlevel=1,
    breaklinks=true,
    colorlinks=true,
    pdftitle={RollPlace: Improving Macro Placement via Monte Carlo Rollout Search},
    pdfauthor={Qi Zhou, Guojun Liu, Guangzhi Qi, Ming Lu, Jiechu Liu, Zhongli Liu, Jianqun Yang, Xingji Li},
    plainpages=false,
    citecolor=blue,
    pdfborder=0 0 0
}

\renewcommand{\IEEEQED}{\IEEEQEDopen}

\begin{document}
	
\title{RollPlace: Improving Macro Placement via Monte Carlo Rollout Search}

\author{
	Qi Zhou, Guojun Liu, Guangzhi Qi, Ming Lu, Jiechu Liu, Zhongli Liu, Jianqun Yang, \IEEEmembership{Member, IEEE}, \\ and Xingji Li, \IEEEmembership{Senior Member, IEEE}
	
	\thanks{This work was supported in part by the National Natural Science Foundation of China (61976071), the Natural Science Foundation of Heilongjiang Province of China (LH2020F012), and the National Key Research and Development Plan Task of China (2021YFF0307903). (\textit{Corresponding author: Guojun Liu.})}
	\thanks{Qi Zhou, Guojun Liu, Guangzhi Qi, Ming Lu, and Jiechu Liu are with the Faculty of Computing, Harbin Institute of Technology, Harbin, 150001, China. (e-mail: hitliu@hit.edu.cn).}
	\thanks{Zhongli Liu, Jianqun Yang, and Xingji Li are with the School of Materials Science and Engineering, Harbin Institute of Technology, Harbin, 150001, China.}
}

\markboth{RollPlace: Improving Macro Placement via Monte Carlo Rollout Search}%
{RollPlace: Improving Macro Placement via Monte Carlo Rollout Search}

\maketitle
\begin{abstract}
	The application of Reinforcement Learning (RL) in Electronic Design Automation (EDA), particularly for chip placement, has attracted considerable attention in recent years. 
	While existing machine learning (ML)-based approaches have achieved notable progress, they predominantly focus on generating optimal layouts in a single attempt, often producing solutions that require subsequent refinement. 
	To address this limitation, we propose RollPlace, a novel and generalized macro placement framework. 
	RollPlace adopts a two-stage optimization strategy: generating initial placement solutions via machine learning methods or heuristic-based strategies,
	and refining these layouts efficiently by adjusting specific macros derived from the initial stage. 
	This strategy circumvents the sequential generation constraints inherent in traditional RL-based placement methods. 
	Furthermore, RollPlace seamlessly integrates Monte Carlo Tree Search (MCTS) to balance exploration and exploitation, and employs a rollout mechanism for efficient local search. 
	Extensive experiments on the ISPD 2005 benchmark demonstrate that RollPlace outperforms state-of-the-art methods. Additionally, end-to-end experimental results based on OpenROAD across 19 benchmarks show that RollPlace excels in multiple metrics.
	The proposed framework offers a robust and scalable solution for addressing the growing complexity of modern chip design challenges.
\end{abstract}


\section{Introduction} \label{Sec:Introduction}

\IEEEPARstart{M}{ACRO} placement, a critical stage in chip design, involves positioning macros of varying dimensions on the chip canvas while satisfying stringent design constraints. The quality of macro placement profoundly impacts the performance metrics of the integrated circuits (ICs), including timing, power consumption, area utilization and so on. However, evaluating the quality of a macro layout typically necessitates the complete placement of all macros, often requiring designers to iteratively refine multiple layouts to achieve an optimal configuration. 
This challenge is exacerbated in modern very-large-scale integration (VLSI) circuits, where the presence of hundreds of macros and tens of thousands of standard cells compounds the complexity of the design. As semiconductor technology advances, this scale is expected to grow further, intensifying demands for scalable placement solutions.
Consequently, the ability to rapidly generate efficient and high-quality layouts has become a critical challenge in chip design, driving the need for innovative solutions to address the escalating complexity of modern ICs.

\begin{figure}[htbp]
	\centering 
	\includegraphics[width=0.49\textwidth]{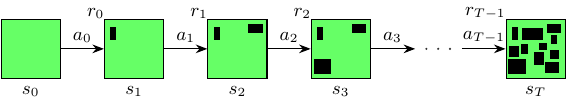} 
	\caption{The general process of placing macros one by one in reinforcement learning. In each state $s_t$, the position of the $t$-th macro is determined according to the decision $a_t$ given by the agent, and the environment gives the reward value $r_t$.} 
	\label{Fig:RL_workflow} 
\end{figure}

In recent years, numerous efforts have been made~\cite{graphplanner, maskplace, chipformer} to apply ML techniques to address the placement problem in EDA. Some of these endeavors have yielded remarkable results, significantly propelling the development and advancement of ML methodologies in the field of chip design.

A common framework adopted by these methods is to model the chip placement task as a Markov Decision Process (MDP)~\cite{graphplanner}. 
As shown in Fig.~\ref{Fig:RL_workflow}, at each step, a macro is positioned on the canvas based on the agent's decision. 
Upon completing the placement of all macros, relevant evaluation metrics, such as wirelength and congestion, are computed based on the macros' positions and the netlist to assess the quality of the placement solution.

In the aforementioned framework, placing the $t$-th macro inherently depends on the decision made at step \(t\!-\!1\), allowing the macro placement results to be represented as a sequence. However, a critical limitation of these approaches lies in their sequential nature: these ML-based approaches often generate the positions of all macros in a single forward pass without additional modifications or adjustments. 
This sequential dependency introduces a significant challenge: the position of a macro with a smaller index in the resulting sequence frequently influences the position of macros with larger indices. 
Generally, a benchmark usually contains at least hundreds of macros, and generating such a long sequence in a single attempt inevitably amplifies and accumulates deviations, ultimately degrading the overall quality of the layout.

Recent attempts have been made to integrate ML-based methods with heuristic techniques to address the instability inherent in sequential macro placement solutions.
Shi et al.~\cite{wiremaskbbo} employed an evolutionary algorithm, named WireMask-EA, to fine-tune the placement results generated by MaskPlace~\cite{maskplace}. 
Nevertheless, WireMask-EA exhibits inherent limitations in optimizing macro placement. Specifically, the algorithm focuses exclusively on refining the single optimal solution provided by MaskPlace, ignoring other potentially viable candidate solutions. This design inherently restricts the optimization space, potentially leading to suboptimal performance. In particular, when the initial placement from MaskPlace produces densely packed macros, the limited flexibility for subsequent adjustments severely constrains the algorithm’s ability to improve the layout.
Another approach, EfficientPlace~\cite{efficientplace}, combines RL techniques with tree search to enhance macro placement. 
This method simultaneously explores better sequences while optimizing the layout strategy through RL. 
However, EfficientPlace adopts a fixed macro ordering strategy, which overlooks the critical influence of macro sequence on placement quality. 
This rigidity can lead to suboptimal outcomes, as unsatisfactory search results in the early stages of the process may propagate errors and adversely affect the subsequent layout strategy. 
Consequently, the pre-determined ordering of macros limits the algorithm's ability to adapt and explore alternative configurations that might yield superior results.


\begin{figure*}[htbp]
	\centering 
	\includegraphics[width=1\textwidth]{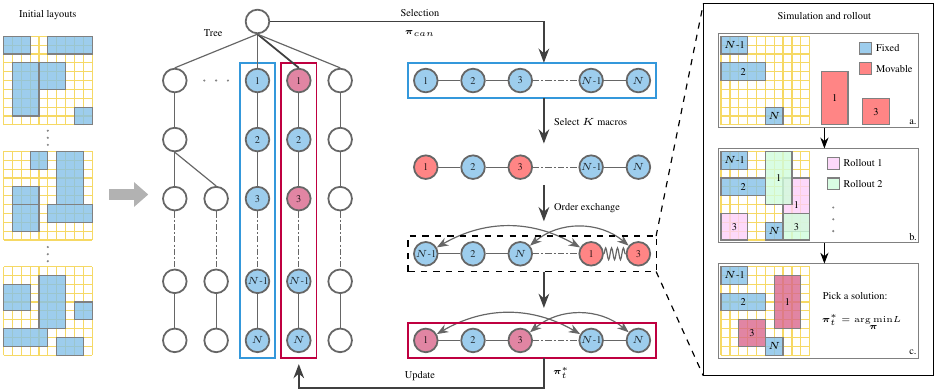} 
	\caption{RollPlace framework. In the first stage, multiple initial layouts are generated and a search tree is constructed using ML methods or others. In the second stage, candidate solutions are refined via a Monte Carlo rollout search, including selection, order exchange, simulation, rollout and update. The right black rectangle illustrates the process of simulation and rollout.} 
	\label{Fig:Framework} 
\end{figure*}


In this paper, we propose RollPlace, a novel macro placement framework that synergistically integrates ML techniques and heuristic functions, as illustrated in Fig.~\ref{Fig:Framework}. 
RollPlace initially employs ML or heuristic layout strategies to generate initialization sequences. This capability enables the efficient generation of candidate sequences for chips that have not been previously encountered, significantly enhancing the framework's generalization ability. 
Subsequently, these initialization sequences are utilized to construct a search tree, which is iteratively refined using MCTS and rollout strategies~\cite{silver2016mastering, silver2017mastering, tesauro1996line}. 
This iterative refinement process allows RollPlace to explore and optimize the solution space effectively, overcoming the instability issues inherent in traditional one-pass ML-based methods.
The key contributions of this work are summarized as follows:
\begin{itemize}
	\item We identify and address a critical limitation in existing ML-based macro placement methods: the instability of one-time sequence generation. To mitigate this issue, we propose a novel framework that combines ML with heuristic algorithms through the construction of a Monte Carlo search tree.
	\item To overcome the challenge of limited or non-existent position adjustment flexibility for macros with higher index values, we introduce a swap-based rollout algorithm. This component seamlessly integrates with the Monte Carlo rollout search, enabling dynamic adjustments to macro positions and significantly improving placement quality. 
	\item Extensive experimental evaluations demonstrate that RollPlace consistently outperforms state-of-the-art methods across multiple benchmarks in macro placement tasks. Furthermore, the framework exhibits robust performance in mixed-size placement tasks, highlighting its versatility and practical applicability in real-world chip design scenarios. The end-to-end evaluation based on OpenROAD~\cite{kahng2021openroad} demonstrates that our algorithm retains significant advantages throughout the complete chip design workflow.
\end{itemize}

\section{Related Work} \label{Sec:RelatedWork}

\subsection{Optimization-based Methods} \label{subSec:OptimizationBasedMethods}
The macro placement problem, fundamentally a combinatorial optimization challenge, can be addressed through various optimization algorithms.
At present, optimization-based placement methods can be broadly categorized into three main types: partitioning-based, simulation annealing-based, and analytical-based approaches.

\textbf{Partition-based Methods}. Partition-based methods~\cite{mincut, recursivebesection} adopt the divide-and-conquer strategy by initially dividing the entire circuit into smaller sub-circuits. 
Specifically, after partitioning, the placement problem for each sub-circuit is addressed independently within its designated sub-region on the chip. Subsequently, these sub-regions are adjusted to suitable positions on the chip. 
However, a significant challenge may arise when dealing with circuits exhibiting diverse topological structures, as partitioning them into relatively independent sub-circuits can be very challenging. 

\textbf{Simulation Annealing-based Methods}. For another perspective, macro placement can be formulated as a Black-Box Optimization (BBO) problem. 
Simulated Annealing (SA) methods~\cite{dragon2000,vashisht2020placementintegratedcircuitsusing, sa3}, also referred to as hill-climbing methods, are widely used iterative heuristic algorithms for solving such BBO problems. These methods generate new solutions through genotype perturbation and phenotype evaluation. 
Furthermore, various solution representation techniques have been proposed, 
including sequence pair (SP)~\cite{sequencepair} , B*-tree~\cite{b*-trees}, and corner block list (CBL)~\cite{CBL}. 
These genotype representations are subsequently translated into specific macro placement configurations for evaluation. 
Recently, the WireMask-BBO framework~\cite{wiremaskbbo} introduced the concept of wire mask~\cite{maskplace}, which enhances the efficiency of BBO algorithms in the genotype space and facilitates integration with multiple BBO algorithms. 

To accurately capture the structural characteristics and dataflow of Register Transfer Level (RTL) designs produced by automatic RTL generators, Kahng et al. proposed a novel macro placement tool named RTL-MP~\cite{RTL-MP}. RTL-MP clusters RTL netlists to reduce problem scale while preserving dataflow and functionality, and places the clusters using simulated annealing. To further enhance the capability of handling large-scale and complex IP block macro placements, Kahng et al. subsequently proposed Hier-RTLMP~\cite{Hier-RTLMP2024}, a hierarchical macro placement framework that leverages the intrinsic hierarchy and dataflow characteristics of RTL designs for multi-scale hierarchical optimization. Notably, Hier-RTLMP has been integrated into the open-source RTL-to-GDSII toolchain OpenROAD~\cite{kahng2021openroad}, thereby strengthening OpenROAD's capacity to deliver reproducible and scalable PPA evaluation for academic research.

\textbf{Analytical-based Methods}.
Analytical-based methods relax the placement task to a mathematical programming problem. 
These approaches transform the optimization objectives into analytical functions of module coordinates. Based on the nature of these analytical functions, analytical-based methods can be broadly classified into two categories: quadratic methods~\cite{linear12,linear13,linear14,linear15,linear16,linear17,linear18} and nonlinear (non-quadratic) methods~\cite{nonlinear6,nonlinear7eplace,nonlinear8replace,nonlinear9dreamplace,nonlinear19,nonlinear20}. Quadratic methods are computationally efficient but often produce lower-quality results. 
In contrast, nonlinear optimization methods provide a more accurate approximation of the cost function, albeit at the expense of significantly higher computational complexity. 
Recently, DREAMPlace~\cite{nonlinear21, nonlinear9dreamplace}, a high-performance analytical-based method, has been developed. 
Built upon the foundations of NTUplace3~\cite{nonlinear6}, ePlace~\cite{nonlinear7eplace} and RePlace~\cite{nonlinear8replace}, DREAMPlace leverages the PyTorch deep learning toolkit to reformulate the analytical placement problem as a neural network training task, thereby significantly accelerating the placement process. 
Further extending the capabilities of DREAMPlace, Agnesina et al.~\cite{AutoDMP} proposed an enhanced framework that incorporates multi-objective Bayesian optimization for macro placement, thereby improving the potential achievable power-performance-area (PPA) by tuning the parameters. Liu et al.~\cite{Xplace} also developed an extremely fast and extensible GPU-accelerated placer called Xplace, which considers operator-level optimization factors. Xplace achieves a 2x speedup compared to DREAMPlace while obtaining higher-quality solutions.

\subsection{Learning-based Methods} \label{Learning-based Methods}

The escalating complexity of modern chip designs, driven by the increasing number of macros, has significantly challenged traditional placement methods. 
In response to these challenges, RL-based approaches have emerged as a promising solution for macro placement, demonstrating remarkable potential in addressing the limitations of conventional techniques.

The foundation of RL-based macro placement was established by Google's pioneering work~\cite{mirhoseini2021graph}, which formulated the macro placement task as a Markov Decision Process. This approach focused on minimizing the Half-Perimeter Wirelength (HPWL) but faced the challenge of sparse reward signals, which hindered efficient learning. To address this limitation, MaskPlace~\cite{maskplace} introduced a novel framework incorporating three specialized masks: view mask, position mask, and wire mask. The wire mask, in particular, enabled the calculation of immediate rewards for macro positions, effectively mitigating the sparse reward problem and enhancing the learning efficiency of RL agents.

Further advancements include PRNet~\cite{cheng2022policy}, which incorporates the overlap area of macros into its reward function as a penalty term. 
This approach effectively reduces the potential issue of overlapping placement that violates design constraints. 

In response to the problem that RL-based methods take a long time to train in unseen circuits, Lai et al.~\cite{chipformer} proposed a new transferable placement method based on transformer architecture~\cite{Ashish2017Attention,gpt,DecisionTransformer}, named ChiPFormer. 
Their approach leveraged offline training with collected placement data and online fine-tuning for unseen chip circuits~\cite{OnlineDecisionTransformer}, effectively reducing the runtime while maintaining high placement quality. 

Despite these advancements, existing RL-based macro placement algorithms often suffer from inefficiency, typically requiring extensive offline expert data or substantial online training. To overcome this limitation, Geng et al.~\cite{efficientplace} proposed EfficientPlace, a sample-efficient framework that combines local policy learning with global tree search algorithms. This innovative approach enables iterative optimization of placements, achieving superior quality with reduced runtime compared to traditional RL-based methods.



\section{Preliminary} \label{Preliminary}

Modern chip design is a highly structured process comprising multiple interdependent stages, including System Specification, Architecture Design, RTL Design, Logic Synthesis, Physical Design, and others~\cite{wang2023circuit}. 
Among these, chip placement constitutes a critical phase within Physical Design, where the primary objective is to optimize the spatial arrangement of movable macros and standard cells on the chip layout. 
This optimization must satisfy a complex set of constraints, including but not limited to overlap avoidance, congestion management, and density distribution control, while simultaneously striving to minimize critical design metrics such as wirelength and power consumption.

Contemporary macro placement benchmarks, which serve as the foundation for evaluating placement algorithms, typically feature hundreds of macros and tens of thousands of standard cells. 
It is noteworthy that these numbers are substantially higher in state-of-the-art industrial chip designs, further exacerbating the computational complexity of the placement problem. 
Since existing ML-based methods place cells sequentially, the resulting design space becomes excessively large, making it computationally intractable for neural networks to handle effectively. 
Consequently, most ML-based approaches focus solely on determining the optimal placement of macros on the layout, while the placement of standard cells is typically resolved using alternative methods, such as the force-directed approach~\cite{mirhoseini2021graph} or advanced tools like DREAMPlace~\cite{nonlinear9dreamplace}.

\subsection{Problem Formulation} \label{subSec:Formulation}

\textbf{Macro placement}. The input to macro placement can be expressed as a hypergraph $H = (V, E)$, where $V$ represents the set of nodes (macros) and $E$ represents the set of edges (netlist), which describes the connection relationships between nodes and the offset of pins.

The macro placement solution can be expressed as a sequence $\{(x_i, y_i)\}_{i=1}^N$ that describes the locations of the macros, where $(x_i, y_i)$ represents the two-dimensional coordinates of the macro $i$ on the physical layout. $N$ is the total number of macros.

To simplify the problem, this paper focuses on minimizing HPWL while disregarding overlap constraints, as presented in the following equation:
\begin{equation}
	\begin{aligned}
		\min & \operatorname{HPWL}(H) \\
		\text { s.t. } & \text { Overlap }(\boldsymbol{x}, \boldsymbol{y}, \boldsymbol{w}, \boldsymbol{h}) = 0
	\end{aligned}
\end{equation}
where $x$ and $y$ denote the positions of macros, $w$ and $h$ represent the width and height of macros.

\textbf{HPWL}. Following established practices in physical design optimization, we employ HPWL as a computationally efficient approximation for estimating interconnect length, as the exact wirelength can only be determined after global routing completion. 
HPWL serves as a fundamental metric for wirelength estimation due to its computational efficiency and strong correlation with actual routing length~\cite{caldwell1998wirelength, chen2006high}. 
The HPWL for a given net is calculated as the sum of the horizontal and vertical spans of its minimum bounding box, encompassing all connected pins. 
Given a specific net $e_{i}$, the HPWL is mathematically defined as:
\begin{equation}
	\begin{aligned}
		\operatorname{HPWL}_{i}  & = \left(\max _{b \in e_i}\left\{x_{b}\right\}-\min _{b \in e_i}\left\{x_{b}\right\}\right) \\
		& + \left(\max _{b \in e_i}\left\{y_{b}\right\}-\min _{b \in e_i}\left\{y_{b}\right\}\right)
	\end{aligned}
\end{equation}
where $x_b$ and $y_b$ represent the two-dimensional coordinates of pin locations, determined by the placement of macros and the relative offsets of pins within net $e_i$. The total HPWL for the complete hypergraph $H$ is then computed as the summation of individual net HPWL across all nets in the design, as expressed in~(\ref{Eq:SumHPWL}):
\begin{equation}\label{Eq:SumHPWL}
	\operatorname{HPWL}(H) = \sum_{e_{i} \in E} \operatorname{HPWL}_i
\end{equation}
%

\subsection{Notations} \label{subSec:Notations}

We define the problem of macro placement $H = (V, E)$ with $N$ macros, where node $i \in \{1, \cdots, N\}$ is represented by $m_i \in V$ and net $i$ is denoted by $e_i \in E$. 
For macro placement, $H$ is a hypergraph which means two nodes may be directly connected by multiple different lines. 
A solution is defined as $\boldsymbol{\pi} = \{ \pi_1, \cdots, \pi_N \}$, where $\pi_i = (x_i, y_i)$ is the coordinate of node $m_i$ in physical layout. 
Similar to the previous work, the entire canvas can be divided evenly into $M \times M$ grids (where M represents the number of divisions along each edge of the canvas, and is a positive integer), with each grid representing a coordinate position. 
Therefore, all coordinate values of the solution are selected from discrete $M \times M $ grids.

We place macros on the canvas step by step, which can be represented as a sequential decision model. For notation consistency, we define the empty sequence $\pi_{1:0} = \emptyset$ (i.e., no previous decisions) when $t=1$. In our framework, the heuristic function can generate a stochastic policy $p(\boldsymbol{\pi} | H)$ to determine a solution $\boldsymbol{\pi}$ given a problem instance $H$ and previous solutions $ \pi_{1: t-1}$:
\begin{equation}
	p(\boldsymbol{\pi} | H) = \prod_{t = 1}^{N} p\left(\pi_{t} | H, \pi_{1: t-1}\right).
\end{equation}

In the context of macro placement, once the position of the node preceding step 
$i$ is determined, their order can be arbitrarily rearranged without affecting the state of the current node $m_i$ or altering the placement policy 
$p$ for the remaining nodes. As a result, the order of the preceding nodes is irrelevant in our framework, whereas their coordinates play a critical role.


\section{Method} \label{Sec:Method}

\subsection{Motivation} \label{subSec:Motivation}

\textbf{One-Time Generation}. 
Modern ML-based macro placement methods are predominantly implemented with RL algorithms, where the placement problem is formulated as a Markov Decision Process. In RL-based framework, macros are placed sequentially on the canvas, with each placement decision influencing subsequent states and actions. 
Specifically, the position of each newly placed macro becomes part of the state space that determines the placement of subsequent macros. 
This sequential dependency creates a deterministic generation process where the final placement configuration is fully determined once the last macro is placed. 
We characterize this approach as one-time generation, which inherently limits the flexibility to refine or adjust placements after the initial generation process.
This one-time generation paradigm stands in stark contrast to traditional manual placement methodologies. 
In manual macro placement, designers maintain complete flexibility to select and reposition any macro at any suitable position on the canvas, without adhering to a predefined order. 
This iterative refinement capability allows for continuous optimization and adjustment of placement configurations, enabling designers to achieve superior placement quality. Consequently, it is essential to fine-tune the results of one-time generation to achieve better placement outcomes.

\textbf{Fixed Macro Ordering}. 
Another limitation of the sequential decision-generation approach is that macros for a given benchmark are typically placed in a fixed order, which can lead to the issue of error accumulation. 
If an early macro placement is suboptimal, subsequent decisions become constrained by this initial error, potentially propagating and amplifying the degradation of placement quality. 
Consequently, in a step by step generation framework, dynamically adjusting the macro placement order can effectively mitigate the error accumulation problem.

\subsection{Framework} \label{subSec:Framework}
To address these challenges, we propose a novel framework that integrates ML with Monte Carlo tree search to resolve the previous issues. 
Additionally, the framework employs heuristic search and a non-deterministic macro ordering strategy to overcome the limitations associated with fixed macro placement sequences, as illustrated in Fig.~\ref{Fig:Framework}. 
First, a set of initial solutions is generated with RL approach or heuristic-based strategies, to construct a search tree in which each solution represents a distinct branch.
Subsequently, we select an appropriate sequence $\{(x_i, y_i)\}_{i=1}^N$ as a candidate solution $\boldsymbol{\pi}_{can}$ from the tree and perform a local search on $\boldsymbol{\pi}_{can}$ in a swappable order. Once a new solution $\boldsymbol{\pi}_{new}$ is generated, the positional order is adjusted to align with $\boldsymbol{\pi}_{can}$. 
Finally, the HPWL~\cite{caldwell1998wirelength, kahng2003accurate} of $\boldsymbol{\pi}_{new}$ is computed, and the search tree is updated accordingly. 
This iterative cycle—selection, order exchange, simulation, rollout, and update—continues until reaching a predefined number of iterations.

The proposed algorithm is structured into two distinct stages.
In the first stage, a set of initial solutions $\{\boldsymbol{\pi}_i\}_{i=1}^P$ is generated, and a search tree is constructed based on these solutions. 
In the second stage, the one-time generated solutions are further refined via a Monte Carlo rollout search. 
The flow of the second stage is summarized in Algorithm~\ref{Alg:RollPlace}. Code is publicly available at https://github.com/smartbird66/RollPlace.

\subsection{Initialize Placement} \label{subSec:init}

The transferable ML-based methods or fast placers are recommended to initialize placement in our framework. These approaches significantly reduce the time required for chip placement. Taking ChiPFormer~\cite{chipformer} as an example, unlike other RL placement methods that rely on convolutional neural networks (CNNs), a sequential decision-making module $f_\theta$ with parameter $\theta$ was trained on a chip placement dataset, followed by an offline fine-tuning for the target chip. 
For different target chips, only a simple fine-tune of the pre-trained model is required, eliminating the need to train a neural network from scratch.

\subsection{Monte Carlo Rollout Search} \label{subSec:MCTS}
Our RollPlace framework is depicted in Fig.~\ref{Fig:Framework}. 
In the search tree, each solution is represented as a fixed sequence $\{\pi_i\}_{i=1}^N$. Since fixed-order macro placement can degrade placement quality, we propose a rollout strategy to perform randomized local optimization.
This strategy randomly swaps the placement order of selected macros and employs a stochastic approach to locally optimize the sequence, thereby addressing the fixed order limitation inherent in existing methods. 
Our method further integrates a heuristic search strategy based on position mask and wire mask, as introduced in~\cite{maskplace}. 
Specific details of the strategy will be elaborated in section~\ref{SubSec:HF}.

\begin{figure}[t]
	
	\def \wThreefig {0.30\columnwidth}
	
	\centering 
	\begin{minipage}{\columnwidth}
		\centering
		\begin{minipage}{\wThreefig}
			\subfloat[$\boldsymbol{\pi}$]{ \includegraphics[width=\textwidth]{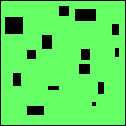} }
		\end{minipage}
		\hfill
		\begin{minipage}{\wThreefig}
			\subfloat[$\boldsymbol{\pi}_{sub}$]{ \includegraphics[width=\textwidth]{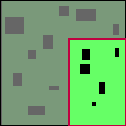} }
		\end{minipage}
		\hfill
		\begin{minipage}{\wThreefig}
			\subfloat[$\boldsymbol{\pi}-\boldsymbol{\pi}_{sub}$]{ \includegraphics[width=\textwidth]{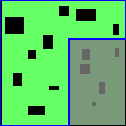} }
		\end{minipage}
	\end{minipage}
	\caption{Macro adjustment. (a) $\boldsymbol{\pi}$ is the original solution, (b) $\boldsymbol{\pi}_{sub}$ is the subsequence which are going to be adjust, and (c) $\boldsymbol{\pi}-\boldsymbol{\pi}_{sub}$ keeps the original position.} 
	\label{Fig:MacroAdjust} 
\end{figure}

The flexibility to modify the macro placement order stems from the following property of macro sequences: once all macros are placed under constraints, the position of any one or more macros can be adjusted, provided that the new positions still satisfy the constraints. Such adjustments result in a valid solution. 
For instance, as illustrated in Fig.~\ref{Fig:MacroAdjust}, consider a solution $\boldsymbol{\pi}$ where all macros satisfy the constraint $overlap = 0$. 
A subsequence $\boldsymbol{\pi}_{sub}$ is sampled from $\boldsymbol{\pi}$, and both $\boldsymbol{\pi}_{sub}$ and $\boldsymbol{\pi}-\boldsymbol{\pi}_{sub}$ satisfy the constraint. By fixing $\boldsymbol{\pi}-\boldsymbol{\pi}_{sub}$ and adjusting $\boldsymbol{\pi}_{sub}$ according to the policy $p$ while maintaining constraint satisfaction, a new sequence $\boldsymbol{\pi}-\boldsymbol{\pi}_{sub} + \boldsymbol{\pi}_{sub}'$ is obtained. 
This new sequence remains a valid solution as it still satisfies the constraint.
This process is equivalent to first placing the macros of $\boldsymbol{\pi}-\boldsymbol{\pi}_{sub}$ in their existing positions, and then placing the macros of $\boldsymbol{\pi}_{sub}$. 
Consequently, randomly swapping macros based on this property does not affect the final placement outcome.


The following section provides a detailed description of the Monte Carlo rollout search.

\textbf{Selection}. 
In the selection phase, identifying an appropriate candidate solution $\boldsymbol{\pi}_{can}$ is critical for the effective execution of subsequent search operations. Starting from the root, successive child nodes are selected until a leaf node is reached. The root node is a virtual node, regarded as layer $0$, while the remaining nodes are organized into layers $1$ through $N$ from top to bottom. Each node in layer $i$ corresponds to a policy $\pi_i$. A complete solution is represented by a full path from layer $1$ to layer $N$. To construct a candidate solution, nodes are sampled sequentially from top to bottom.

Each node $a$ in the search tree maintains three statistics:  
(1) the visit count of the node, denoted as $n_a$;  
(2) the cumulative reward of all solutions containing node $a$, denoted as $L_a$;  
(3) the minimum reward value among all solutions containing node $a$, denoted as $L_{a}^{min}$.  

To balance exploration and exploitation, we adopt the Upper Confidence Bound applied to Trees (UCT) strategy~\cite{kocsis2006bandit}. At each step $i$, the node $a$ in layer $i$ is selected by maximizing the following confidence bound, derived from the Chernoff–Hoeffding inequality:

\begin{equation} \label{Eq:Chernoff-Hoeffding-bound}
	a_{can} = \mathop{\arg\max}\limits_{a\in A} \left ( \bar{v}_{a}+C \sqrt{\frac{\log \sum_{b \in A} n_{b}}{n_{a}}} \right ) .
\end{equation}
where $C$ is a hyperparameter controlling the trade-off between exploration and exploitation; $A$ denotes the set of sibling nodes sharing the same parent as node $a$ at layer $i$; $\bar{v}_a$ is the estimated value of node $a$; $n_a$ is the visit count of node $a$; and $n_b$ is the visit count of a sibling node $b \in A$.  

The value $\bar{v}_a$ is defined as:  
\begin{equation}
	\bar{v}_{a} = -( \, L_{a}^{min} + \mathbb{E}(L \mid a) \, ) / 2
\end{equation}

where $L_{a}^{min}$ denotes the minimum HPWL value among all solutions containing node $a$, and $\mathbb{E}(L\!\mid\!a)$ represents the expected HPWL of solutions containing node $a$. 

As illustrated in Fig.~\ref{Fig:Framework}, the branch highlighted in the blue rectangle corresponds to the selected candidate solution $\boldsymbol{\pi}_{can}$. Once a complete solution $\boldsymbol{\pi}_{can}$ is sampled, order exchange is performed.

\textbf{Order Exchange}. 
To resolve the issue of fixed macro ordering, an order exchange operation is performed prior to the simulation and rollout. For the selected candidate solution $\boldsymbol{\pi}_{can}$, represented as a sequence $\{ \pi_{i}\}_{i=1}^N$, $K$ distinct indices are generated by random uniform sampling from the set $\{ 1, \cdots, N \}$.

The sampled $K$ indices are utilized to determine the macros to be swapped. These macros are subsequently relocated to the terminal position of the predefined sequence.
This operation splits the solution into two complementary subsquence: $\boldsymbol{\pi}_{sub}$ (the sampled macros) and $\boldsymbol{\pi}_{can}-\boldsymbol{\pi}_{sub}$ (the remaining macros). Subsequently, the subsequence $\boldsymbol{\pi}_{sub}$ undergoes local search with rollouts, while $\boldsymbol{\pi}_{can}-\boldsymbol{\pi}_{sub}$ remains unchanged.

Consider an illustrative case with $K=2$, assuming that the randomly sampled indices are $\{1,3\}$ (see Fig.~\ref{Fig:Framework}). 
The macros corresponding to indices $1$ and $3$ are swapped with the macros at indices $N-1$ and $N$, respectively, resulting in a new sequence order $\{N-1, 2, N, 4, \cdots, N-2, 1, 3\}$. The two complementary subsquence are $\boldsymbol{\pi}_{sub} = \{\pi_1, \pi_3\}$ and $\boldsymbol{\pi}_{can}-\boldsymbol{\pi}_{sub} = \{\pi_{N-1}, \pi_{2}, \pi_{N}, \pi_{4}, \cdots, \pi_{N-2}\}$.

\textbf{Simulation}. 
The purpose of the simulation in this work is to replicate the state where the sequence $\boldsymbol{\pi}_{can}-\boldsymbol{\pi}_{sub}$ is placed. 
Specifically, the macros are placed on the canvas in the order defined by indices $\{N-1, 2, N, 4, \cdots, N-2\}$, following the placement strategy derived from $\boldsymbol{\pi}_{can}-\boldsymbol{\pi}_{sub}$. 
Notably, the placement process does not require explicit constraint calculations, as the selected solution $\boldsymbol{\pi}_{can}$ is inherently valid and satisfies all necessary constraints. As illustrated in Fig.~\ref{Fig:Framework}-a, all macros except for cells 1 and 3 are directly placed at their designated positions according to the given strategy. Macros 1 and 3, however, are reserved for multiple rollouts in the subsequent stage.

\textbf{Rollout}. 
The rollout phase optimizes the position of macros in subsequence $\boldsymbol{\pi}_{sub}$ based on the simulation result and a heuristic strategy.

During the rollout process, the macros in subsequence $\boldsymbol{\pi}_{sub}$ are placed sequentially on the canvas, guided by simulation results, while ensuring that no overlaps are violated. 
Once all macros in $\boldsymbol{\pi}_{sub}$ are placed, a new sequence $\{\pi_{N-1}, \pi_{2}, \pi_{N}, \pi_{4}, \cdots, \pi_{N-2}, \pi_1', \pi_3'\}$ is acquired, which is slightly modified compared to the original sequence $\boldsymbol{\pi}_{can}$. 
Due to the interchangeable nature of macro placement, the new sequence can be equivalently expressed as $\boldsymbol{\pi}' = \{\pi_1', \pi_{2}, \pi_{3}', \pi_{4}, \cdots, \pi_{N-2}, \pi_{N-1}, \pi_{N}\}$. 
Then, the HPWL $L'$ of the new solution $\boldsymbol{\pi}'$ is computed to evaluate its quality. 

To accelerate the local search process, multiple rollouts $\{\boldsymbol{\pi}_i\}_{i=1}^R$ can be performed based on a single simulation.
Finally, the optimal solution $\boldsymbol{\pi}_t^{*}$ with the best HPWL is selected, ensuring that $L_t^* = \min\{L_1, \cdots, L_R\}$. Fig.~\ref{Fig:Framework}-b and~\ref{Fig:Framework}-c illustrate the detailed procedure of multiple rollouts. The selection rule is formally defined as:  
\begin{equation} \label{Eq:SelectCan}
	\boldsymbol{\pi}_t^* = \underset{\boldsymbol{\pi}_i \in \{\boldsymbol{\pi}_i\}_{i=1}^R }{\arg\min} \, L(\boldsymbol{\pi}_i)
\end{equation}
\textbf{Update}. 
In this process, we update the search tree from top to bottom. 
If the node corresponding to $\boldsymbol{\pi}_t^{*}$ already exists in the tree, the properties of the node need to be updated. 
If the corresponding node does not exist in the search tree, then a new node is constructed and initialized. Its parent node is the node corresponding to its previous index. The node is updated as follows:
\begin{equation}
	\begin{aligned}
		& n_a = n_a + 1 \\
		& L_{a} = L_{a} + L_{t}^* \\
		& L_{a}^{min} = \min (L_{a}^{min}, L_{t}^*) .
	\end{aligned}
\end{equation}
where $n_a$ represents the visit count of node $a$, $L_{a}$ denotes the cumulative HPWL and $L_{a}^{min}$ tracks the minimum observed HPWL.


\begin{algorithm}[htbp]
	\renewcommand{\algorithmicrequire}{\textbf{Input:}}
	\renewcommand{\algorithmicensure}{\textbf{Output:}}
	\caption{RollPlace Algorithm}
	\label{Alg:RollPlace}
	\begin{algorithmic}[1]
		\REQUIRE Circuit macros $V$, Circuit netlist $E$, initialized tree $\mathcal{T}$, search round $T$, sampling count $K$, rollout count $R$          
		\ENSURE Optimized placement solution $\boldsymbol{\pi}^{*}$  
		\FOR{$t \gets 1$ \TO $T$}
		\STATE Sample a candidate solution $\boldsymbol{\pi}_{can} \sim \mathcal{T}$ node by node with (\ref{Eq:Chernoff-Hoeffding-bound}).
		\STATE $K$ distinct indices are generated by random uniform sampling from the set $\{ 1, \cdots ,N \}$.
		\STATE Exchange ordering of the last $K$ macros with the selected $K$ macros.
		\FOR{$j \gets 1$ \TO $N - K$}
		\STATE Keep the macro in the indice $j$ of new sequence order with its original position.
		\ENDFOR
		\FOR{$r \gets 1$ \TO $R$}{
			\FOR{$j \gets N - K + 1$ \TO $N$}
			\STATE Place the macro in the indice $j$ of new sequence with policy from heuristic function $F_{j}$.
			\ENDFOR
			\STATE Calculate HPWL $L_r$ and record new solution $\boldsymbol{\pi}_r$.
		}
		\ENDFOR
		\STATE Select the best solution $\boldsymbol{\pi}_t^{*}$ according to~(\ref{Eq:SelectCan}). 
		\STATE Update tree $\mathcal{T}$ with the best solution $\boldsymbol{\pi}_t^{*}$.
		\ENDFOR
		\RETURN The best placement solution $\boldsymbol{\pi}^{*}$ in $\mathcal{T}$
	\end{algorithmic}
\end{algorithm}


\subsection{Heuristic Function}
\label{SubSec:HF}

Consider an ordered sequence $\{m_1, m_2, \cdots, m_N\}$, where the positions of the first $i-1$ macros $\{m_1, m_2, \cdots, m_{i-1}\}$ have already been determined on the placement canvas. 
The heuristic function $F_{i}$ is computed as the Hadamard product of position mask $P_{i}$ and wire mask $W_{i}$, i.e., by $F_{i} = P_i \odot W_i$. Both $P_i$ and $W_i$ are matrices of size $M \times M$ whose values are dynamically generated based on the current placement result of $\{m_1, \cdots, m_{i-1}\}$.
The placement policy for macro $m_i$ is defined as a uniform probability distribution over all coordinates $(x,y)$ that yield the minimal values in $F_i$.
During the rollout process, this policy is applied iteratively through sequential sampling until all $N$ macros are placed on the grid. 

WireMask-EA~\cite{wiremaskbbo} adopts a greedy strategy by selecting the nearest position within the candidate grid.
In contrast, our approach employs a uniform probability distribution to explore a broader range of possibilities, thereby optimizing for better HPWLs. 
The following subsections provide a brief description of $P_{i}$ and $W_{i}$. 
For a detailed description, please refer to~\cite{maskplace}.

\textbf{Position Mask}. 
$P_{i}$ identifies legal grid positions for macro $m_i$, ensuring compliance with overlap constraints and canvas boundary limits. 
It is represented as a binary 2D grid $\{1,\infty\}^{M \times M}$, where $\infty$ indicates an illegal position and $1$ denotes a legal position.


\begin{table*}[hbp]
	\renewcommand{\arraystretch}{1.2}
	\centering
	\caption{Comparisons of HPWL ($\times10^5$). Lower HPWL values indicate better placement quality. Results for DeepPlace are sourced from~\cite{efficientplace}, while all other methods are evaluated by executing their respective open-source implementations with five distinct random seeds to ensure statistical reliability. For our proposed method, the minimum HPWL values achieved at $500$ and $1000$ optimization steps are recorded. Performance metrics are reported as mean values with standard deviations (mean$\pm$std). The best-performing results are highlighted in \textbf{bold}, while the second-best results are marked in \textcolor{brown}{brown}.}
	\label{TableMainResult}
	\resizebox{\textwidth}{!}{
		\begin{tabular}{c c c c c c c c c} 
			\toprule
			Method & adaptec1 & adaptec2 & adaptec3 & adaptec4 & bigblue1 & bigblue2 & bigblue3 & bigblue4  \\
			\midrule
			DREAMPlace & $13.80\pm0.57$ & $40.57\pm3.06$ & $62.34\pm0.61$ & $353.95\pm43.76$ & $7.38\pm1.48$ & $77.55\pm9.63$ & $86.53\pm3.56$ & $112.95\pm4.03$ \\
			DeepPlace & $19.91\pm2.13$ & $203.51\pm6.27$ & $347.16\pm4.32$ & $311.86\pm56.74$ & $23.33\pm3.65$ & $11.38\pm0.20$ & $430.48\pm12.18$ & $433.90\pm5.26$ \\
			MaskPlace & $7.56\pm0.20$ & $71.13\pm3.93$ & $106.00\pm4.04$ & $81.56\pm3.34$ & $3.16\pm0.13$ & $10.69\pm0.88$ & $82.55\pm6.56$ & $130.15\pm10.94$ \\
			ChiPFormer & $7.40\pm0.09$ & $72.32\pm5.41$ & $80.05\pm0.87$ & $68.44\pm1.29$ &$3.08\pm0.01$ & $7.24\pm0.37$ & $89.36\pm2.10$ & $94.90\pm2.50$\\
			WireMask-EA & $6.23\pm0.09$ & $60.33\pm3.67$ & $63.04\pm1.77$ & $64.99\pm1.17$ & $2.17\pm0.02$ & $6.25\pm0.24$ & $66.79\pm5.26$ & $87.81\pm3.30$  \\
			EfficientPlace & $5.99\pm0.06$ & $44.10\pm1.64$ & $60.29\pm1.73$ & $\boldsymbol{59.46\pm1.60}$ & $2.41\pm0.03$ & $\boldsymbol{4.88\pm0.97}$ & $77.90\pm11.33$ & $88.01\pm2.71$  \\
			RollPlace & $\textcolor{brown}{5.98\pm0.11}$ & $\textcolor{brown}{39.59\pm4.67}$ & $\textcolor{brown}{58.03\pm3.27}$ & $67.19\pm1.87$ &  $\textcolor{brown}{2.05\pm0.04}$ & $5.09\pm0.14$ & $\textcolor{brown}{55.94\pm5.79}$ & $\textcolor{brown}{76.26\pm3.26}$ \\
			RollPlace(1k) & $\boldsymbol{5.93\pm0.07}$ & $\boldsymbol{34.83\pm1.81}$ & $\boldsymbol{54.88\pm3.08}$ & $\textcolor{brown}{64.66\pm3.49}$ & $\boldsymbol{2.05\pm0.03}$ & $\textcolor{brown}{5.02\pm0.21}$ & $\boldsymbol{53.56\pm7.31}$ & $\boldsymbol{69.67\pm2.70}$ \\
			
			\bottomrule
		\end{tabular}
	}
\end{table*}


\textbf{Wire Mask}. 
$W_{i}$ quantifies the HPWL increment caused by the placement of macro $m_i$ on the canvas. 
It is represented as a continuous-valued 2D grid $[0,1]^{M \times M}$, where the value in each grid corresponds to the normalized HPWL increase if the macro $m_i$ is placed on the canvas grid.


\section{Experiments} \label{Sec:Experiments}

\subsection{Macro Placement Experimental Results}


\begin{figure*}[htbp]
	
	\def \wFourfig {0.24\textwidth}
	
	\begin{minipage}{\textwidth}
		\centering
		\begin{minipage}{\wFourfig}
			\subfloat[adaptec1]{ \includegraphics[width=\textwidth]{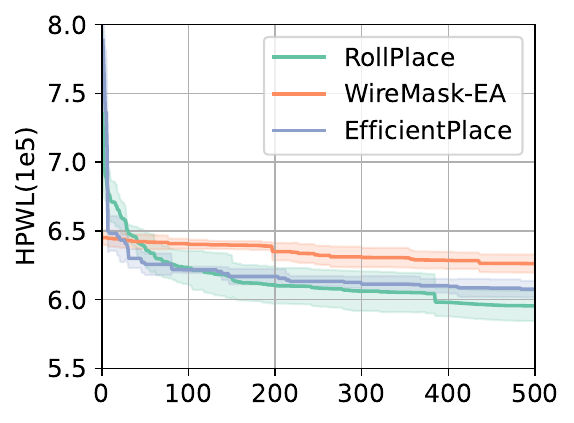} }
		\end{minipage}
		\hfill
		\begin{minipage}{\wFourfig}
			\subfloat[adaptec2]{ \includegraphics[width=\textwidth]{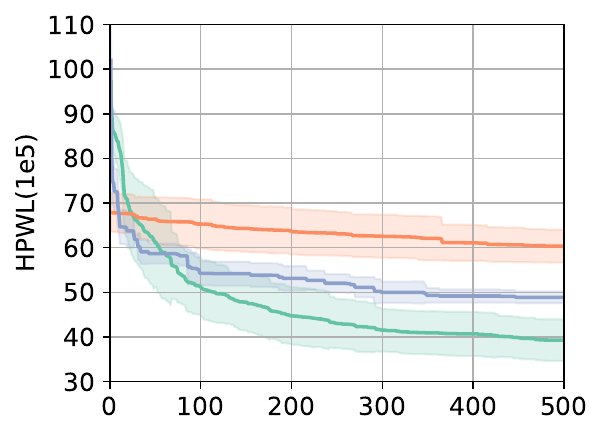} }
		\end{minipage}
		\hfill
		\begin{minipage}{\wFourfig}
			\subfloat[adaptec3]{ \includegraphics[width=\textwidth]{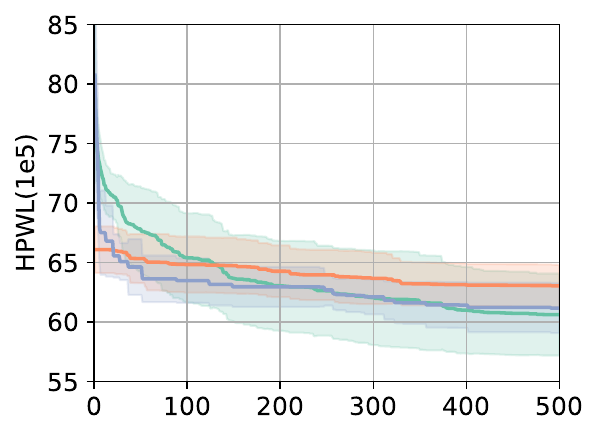} }
		\end{minipage}
		\hfill
		\begin{minipage}{\wFourfig} 
			\subfloat[adaptec4]{ \includegraphics[width=\textwidth]{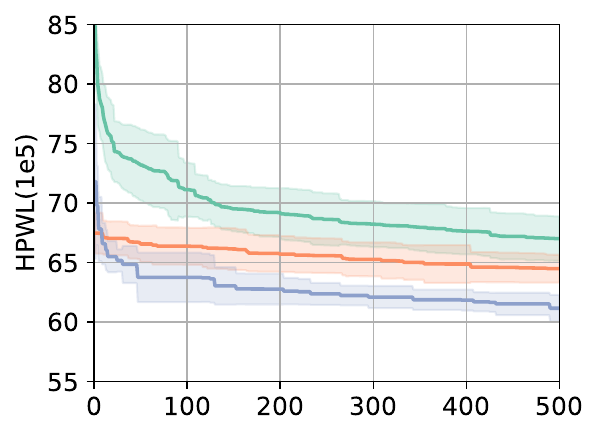} }
		\end{minipage}
	\end{minipage}
	\begin{minipage}{\textwidth}
		\centering
		\begin{minipage}{\wFourfig}
			\subfloat[bigblue1]{ \includegraphics[width=\textwidth]{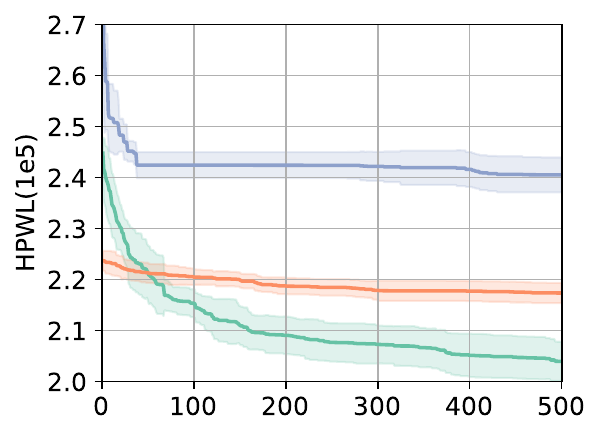} }
		\end{minipage}
		\hfill
		\begin{minipage}{\wFourfig}
			\subfloat[bigblue2]{ \includegraphics[width=\textwidth]{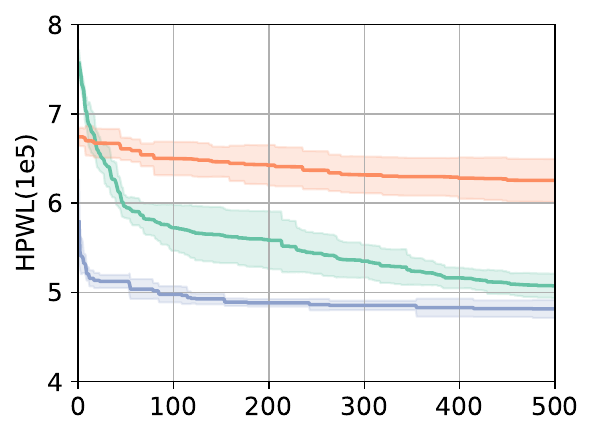} }
		\end{minipage}
		\hfill
		\begin{minipage}{\wFourfig}
			\subfloat[bigblue3]{ \includegraphics[width=\textwidth]{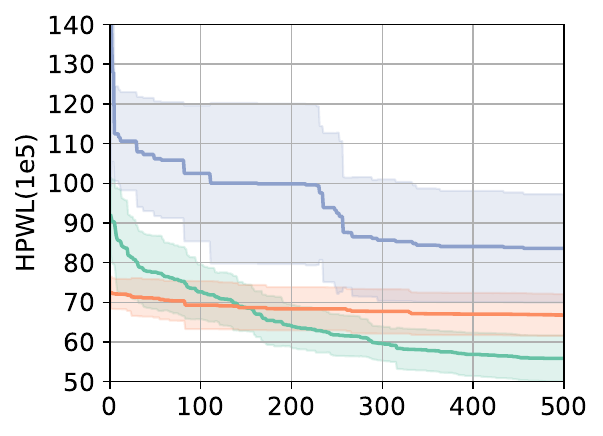} }
		\end{minipage}
		\hfill
		\begin{minipage}{\wFourfig} 
			\subfloat[bigblue4]{ \includegraphics[width=\textwidth]{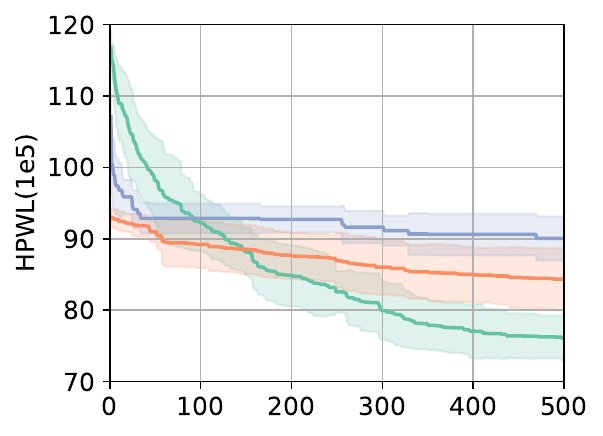} }
		\end{minipage}
	\end{minipage}
	\caption{Convergence analysis of the best HPWL values ($\times10^5$) for RollPlace, EfficientPlace and WireMask-EA across optimization steps. Solid lines denote mean values, while shaded regions represent standard deviation intervals derived from five independent trials with distinct random seeds.}
	\label{Fig:HPWLCurves}
\end{figure*}


We perform a comprehensive evaluation using the ISPD2005 benchmarks~\cite{nam2005ispd2005}, which are widely utilized for macro placement tasks, particularly in recent ML-based research. 
While the original ISPD2005 benchmarks provide predefined fixed positions for all macros, our experimental setup deviates from this configuration by disregarding the fixed macro positions and excluding standard cell placement, in alignment with recent methodologies. 
Our framework is rigorously compared against state-of-the-art methodologies, including DeepPlace~\cite{deepplace}, DREAMPlace~\cite{nonlinear9dreamplace}, MaskPlace~\cite{maskplace}, ChiPFormer~\cite{chipformer}, WireMask-EA~\cite{wiremaskbbo}, and EfficientPlace~\cite{efficientplace}, across all ISPD2005 benchmark circuits, i.e., $adaptec1-adaptec4$ and $bigblue1-bigblue4$. 
Due to the computational complexity caused by the large number of macros in $bigblue2$ (23,084 macros) and $bigblue4$ (8,170 macros), we employ the sampling method proposed in~\cite{efficientplace}, selecting $256$ macros for $bigblue2$ and $1,024$ macros for $bigblue4$, respectively. 

Within these experiments, ChiPFormer~\cite{chipformer} processes a subset of macros during placement, which may result in overlaps. 
However, its metrics are calculated without addressing these overlaps. In contrast, metrics for other methods are calculated by ensuring complete macro placement without any overlap. 

For the RollPlace algorithm, we adopt the following parameter settings: 20 initial placements are used in total, and 10–50 macros are randomly selected in each iteration to execute the rollout strategy, with rollout repeated 30 times. Additionally, we conduct supplementary experiments using Monte Carlo rollout search with two iteration counts (500 and 1000 iterations, respectively) to further validate the effectiveness of RollPlace.

Table~\ref{TableMainResult} summarizes the detailed macro placement results across five distinct random seeds, where the objective is to minimize MacroHPWL. The experimental results demonstrate that RollPlace significantly outperforms recent state-of-the-art approaches on six out of eight benchmark circuits. 
On the other two benchmarks, RollPlace does not achieve the best HPWL, but it consistently attains the second-best values, slightly inferior to EfficientPlace.
Notably, our fine-tuning framework demonstrates an average HPWL reduction of around $30\%$ compared to initial values, highlighting its superior optimization capabilities. 
These results confirm RollPlace's exceptional performance in solving macro placement problems, particularly its ability to surpass recent fixed-order methodologies via adaptive optimization.


\begin{figure*}[htbp]
	
	
	\def \wFourfig {0.12\textwidth}

	\begin{minipage}{\textwidth}
		\includegraphics[width=\wFourfig]{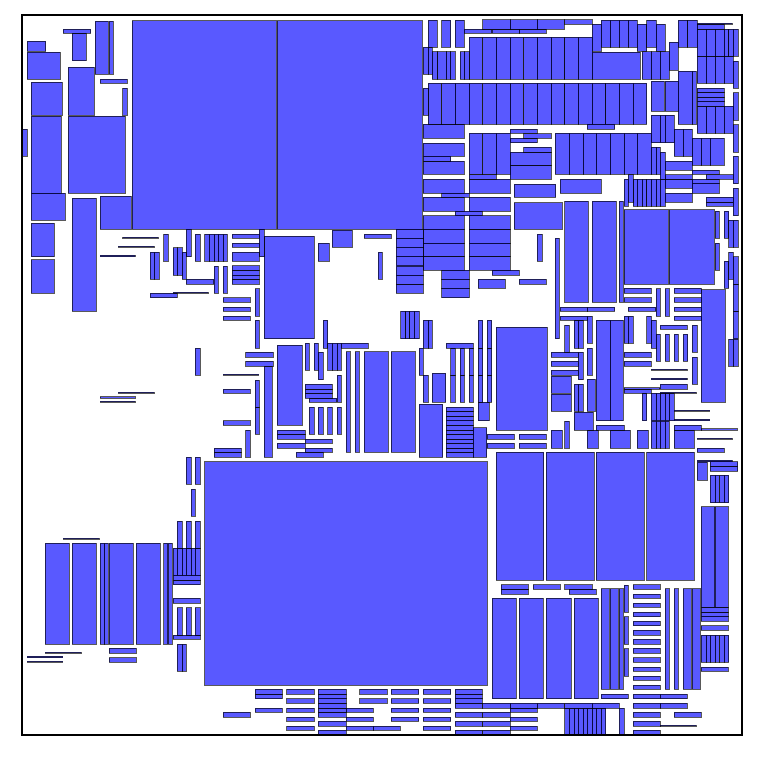}
		\includegraphics[width=\wFourfig]{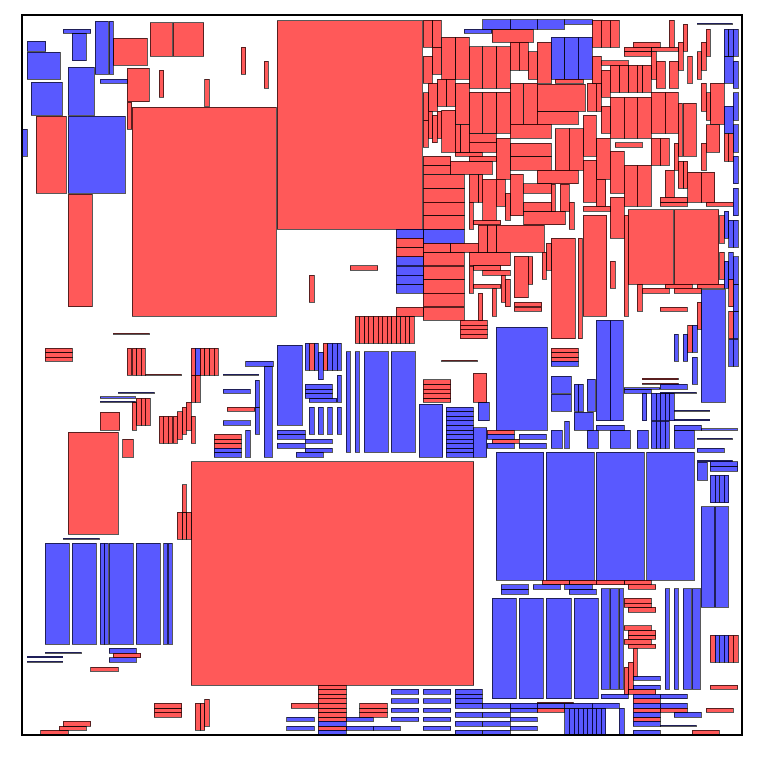}
		\includegraphics[width=\wFourfig]{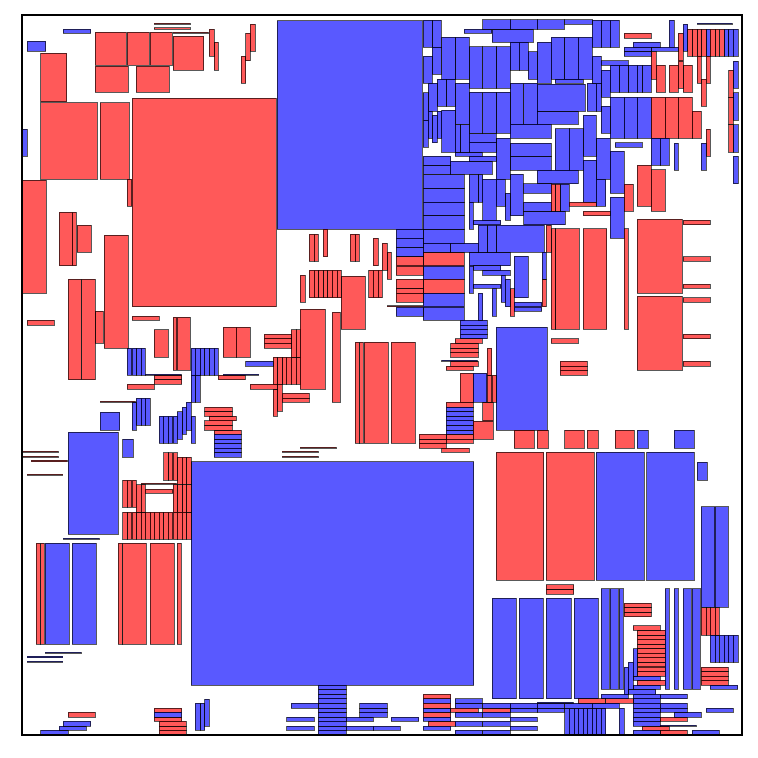}
		\includegraphics[width=\wFourfig]{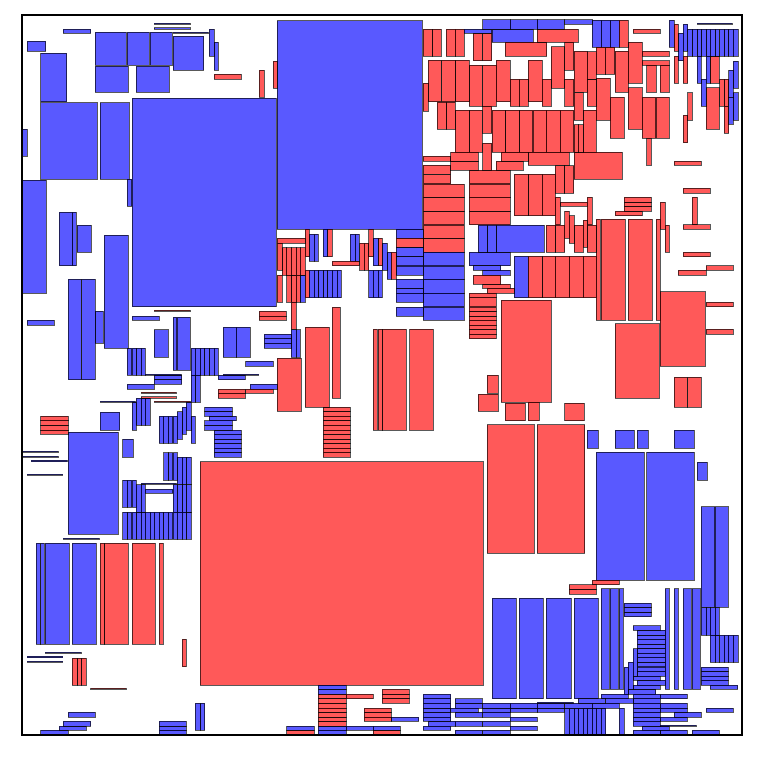}
		\includegraphics[width=\wFourfig]{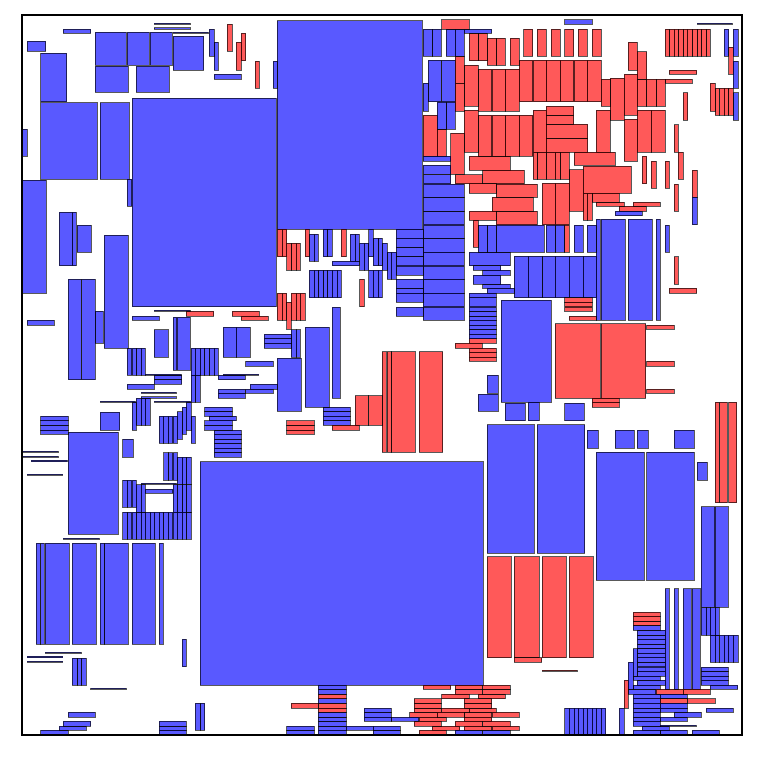}
		\includegraphics[width=\wFourfig]{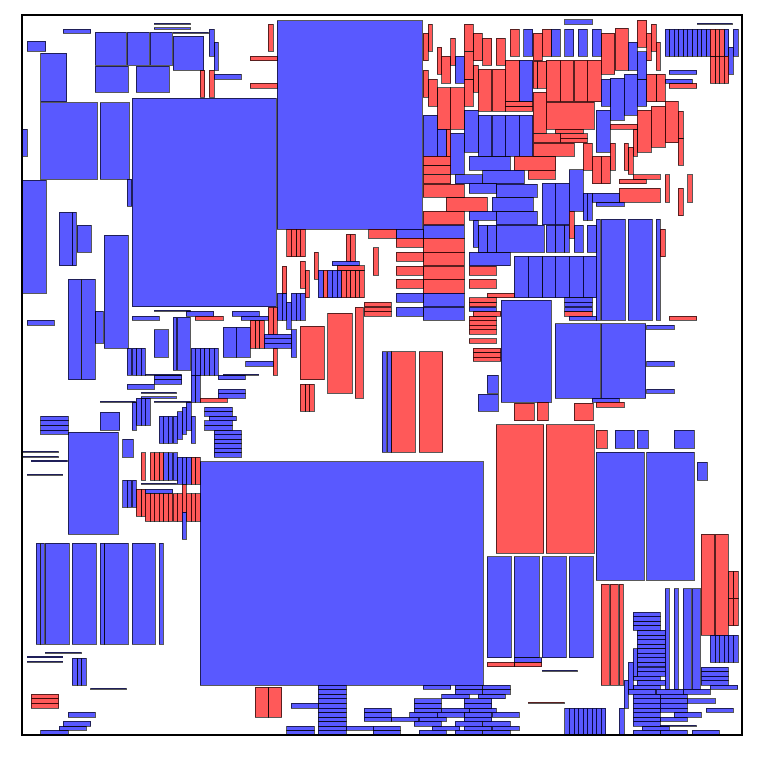}
		\includegraphics[width=\wFourfig]{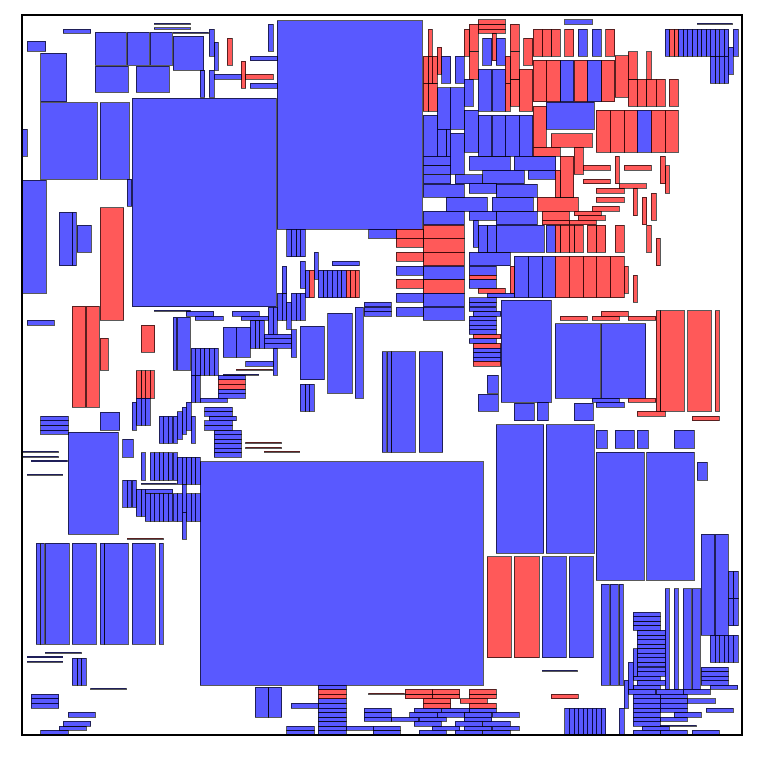}
		\includegraphics[width=\wFourfig]{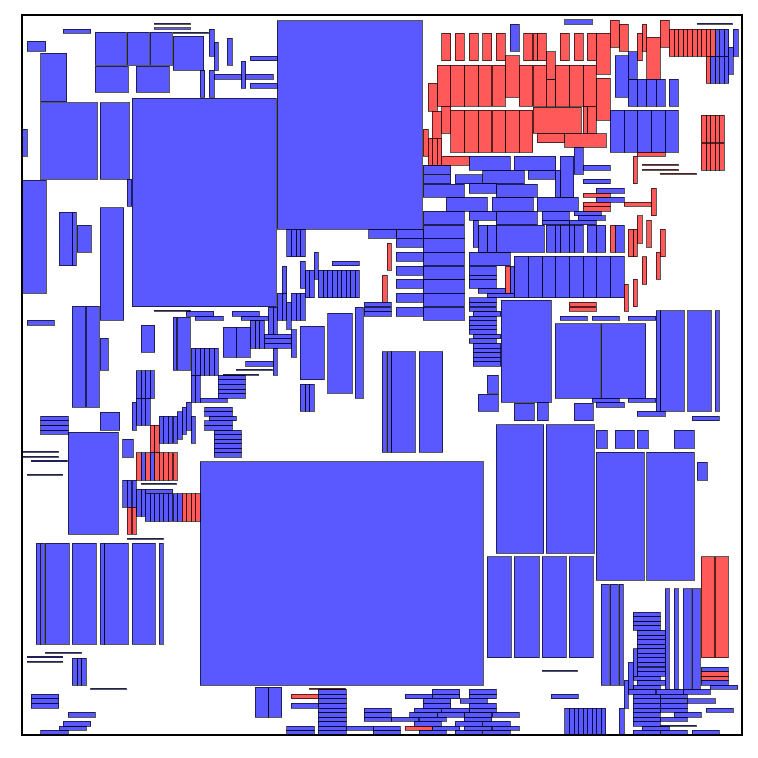}
	\end{minipage}
	\begin{minipage}{\textwidth}
	\end{minipage}
	\caption{Visualization of intermediate placement results during the RollPlace search process on $adaptec2$ benchmark. The red rectangle highlights macros that have shifted positions relative to the prior figure, while the blue rectangle indicates macros maintaining their positions.}
	\label{Fig:InterAdaptec2}
\end{figure*}


Fig.~\ref{Fig:HPWLCurves} illustrates a comparative analysis of the convergence behavior among RollPlace, EfficientPlace, and WireMask-EA, tracking the evolution of best HPWL values across runtime iterations. 
The results show that RollPlace exhibits superior optimization efficiency, characterized by notably faster convergence and more stable performance. 
Additionally, the stability of the final HPWL values across multiple runs highlights the robustness of our framework.

To thoroughly analyze the optimization dynamics of RollPlace, we systematically sampled eight intermediate placement solutions during the search process. 
Fig.~\ref{Fig:InterAdaptec2} visually depicts the progressive refinement of the circuit $adaptec2$ layouts during optimization. The results illustrate that the positions of macros exhibit gradual but significant adjustments as the optimization progresses, particularly in the early and middle search stages. 
These findings confirm the effectiveness of RollPlace in exploring the solution space.

\begin{figure*}[htbp]
	
	
	\def \wSevenfig {0.135\textwidth}
	
	\begin{minipage}{\textwidth}
		\centering
		
		\begin{minipage}{\wSevenfig}
			\subfloat[DREAMPlace \\ $4.007 \times 10^{6}$]{ \includegraphics[width=\textwidth]{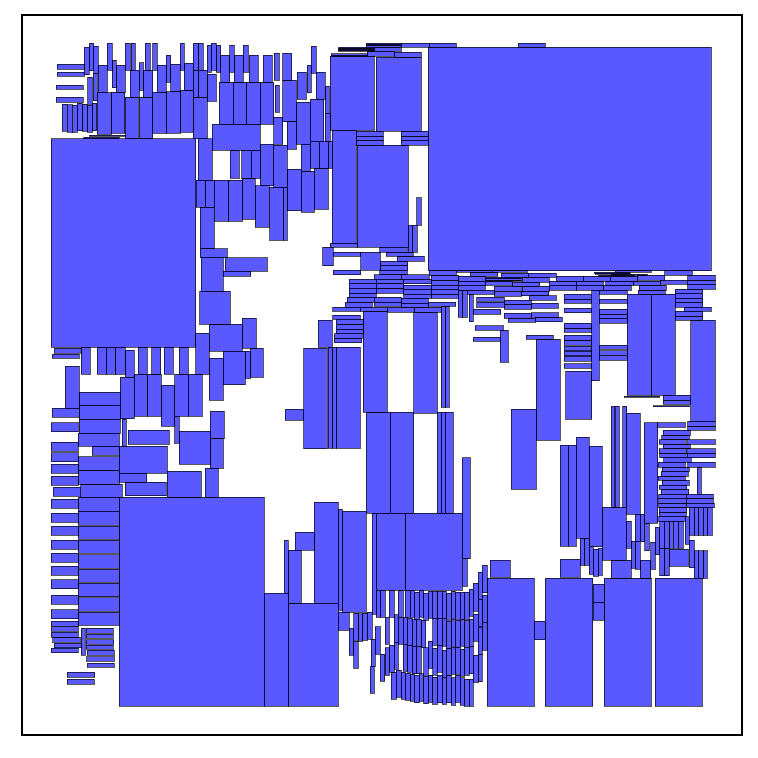} }
		\end{minipage}
		\hfill
		\begin{minipage}{\wSevenfig}
			\subfloat[MaskPlace \\ $7.514 \times 10^{6}$]{ \includegraphics[width=\textwidth]{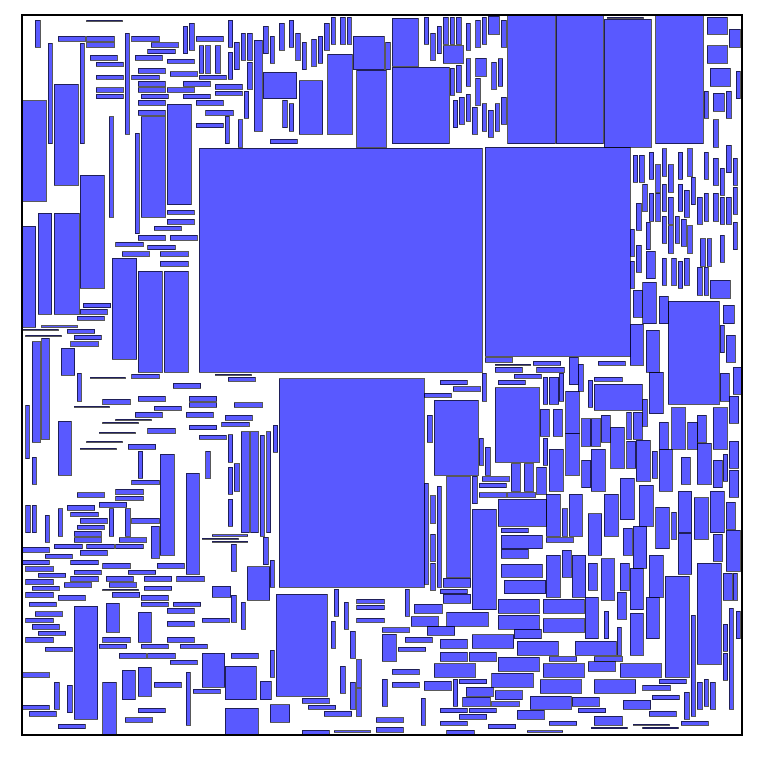} }
		\end{minipage}
		\hfill
		\begin{minipage}{\wSevenfig}
			\subfloat[ChiPFormer \\ $6.501 \times 10^{6}$]{ \includegraphics[width=\textwidth]{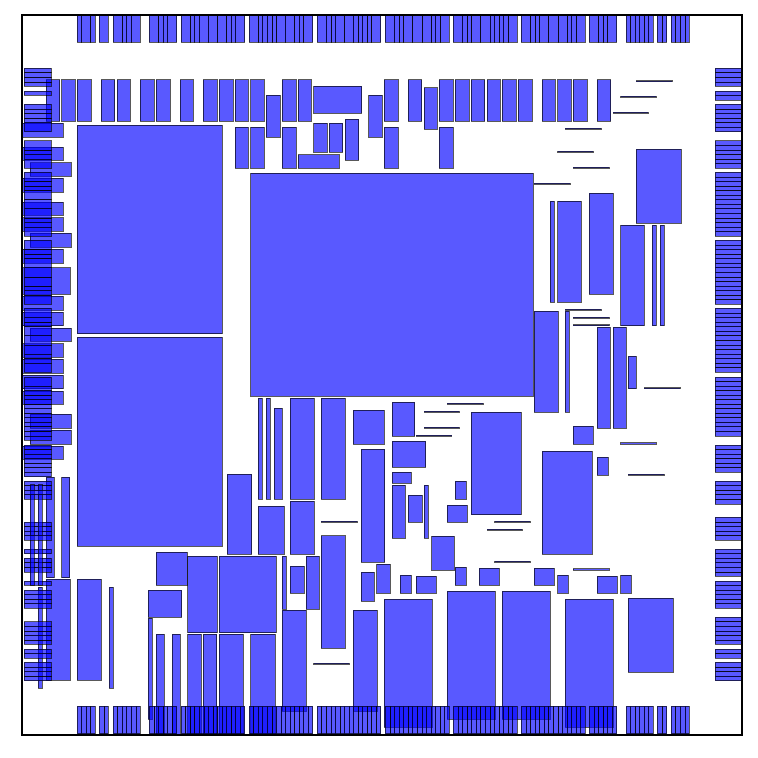} }
		\end{minipage}
		\hfill
		\begin{minipage}{\wSevenfig} 
			\subfloat[WireMask-EA \\ $5.504 \times 10^{6}$]{ \includegraphics[width=\textwidth]{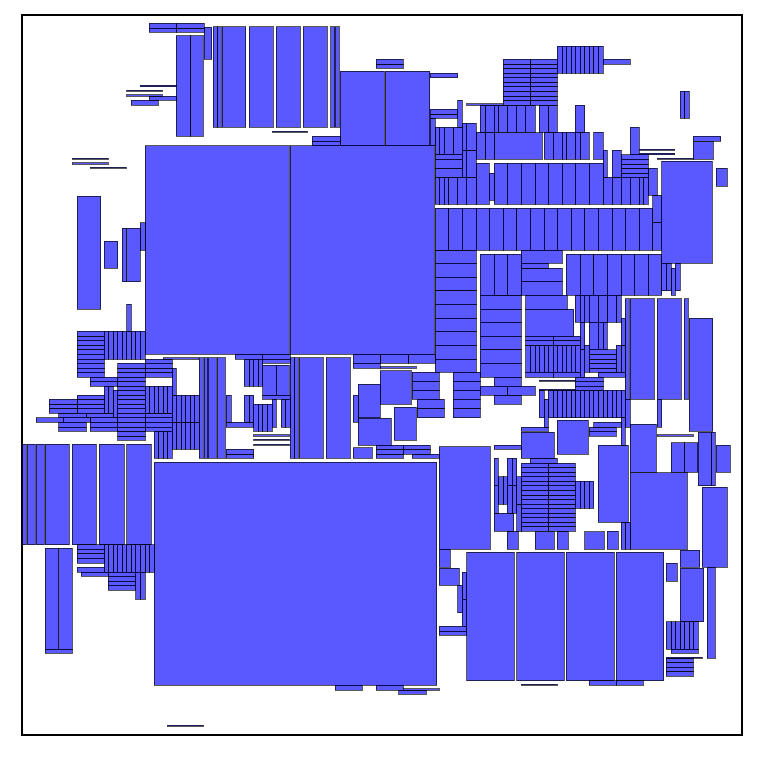} }
		\end{minipage}
		\hfill
		\begin{minipage}{\wSevenfig}
			\subfloat[EfficientPlace \\ $4.326 \times 10^{6}$]{ \includegraphics[width=\textwidth]{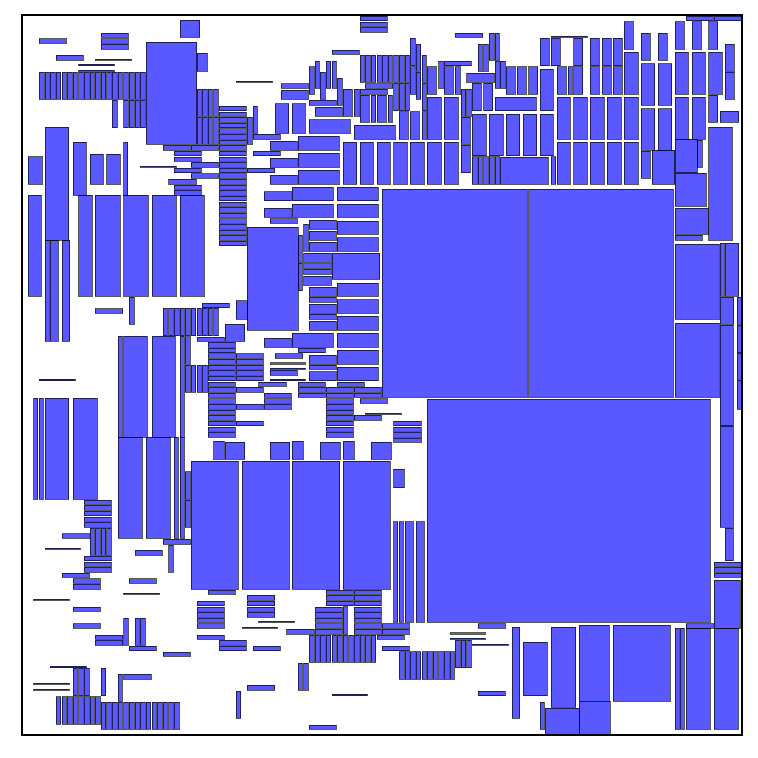} }
		\end{minipage}
		\hfill
		\begin{minipage}{\wSevenfig}
			\subfloat[RollPlace \\ $3.705 \times 10^{6}$]{ \includegraphics[width=\textwidth]{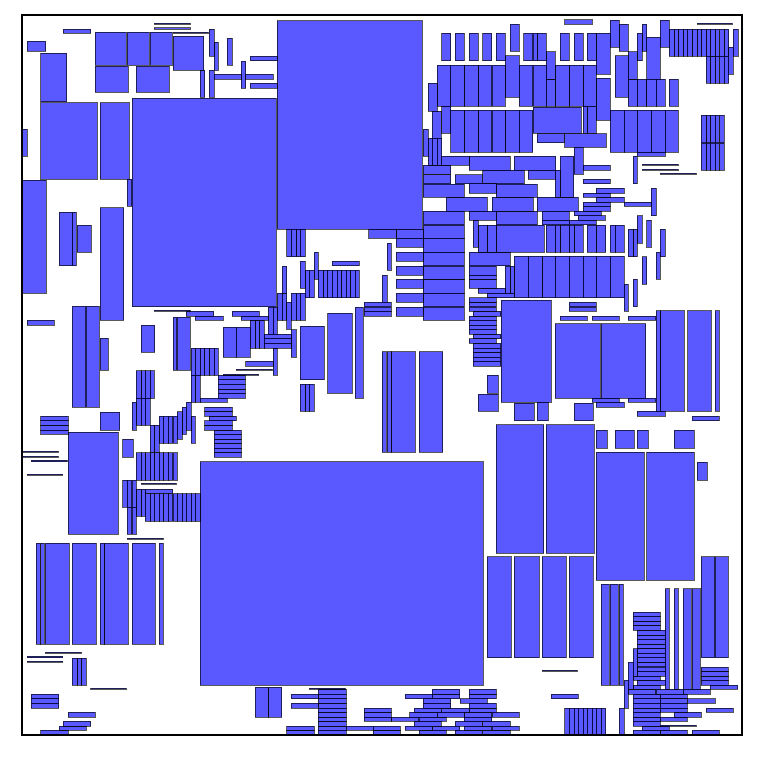} }
		\end{minipage}
		\hfill
		\begin{minipage}{\wSevenfig}
			\subfloat[\textbf{RollPlace (1k) \\ $\boldsymbol{3.449 \times 10^{6}}$}]{ \includegraphics[width=\textwidth]{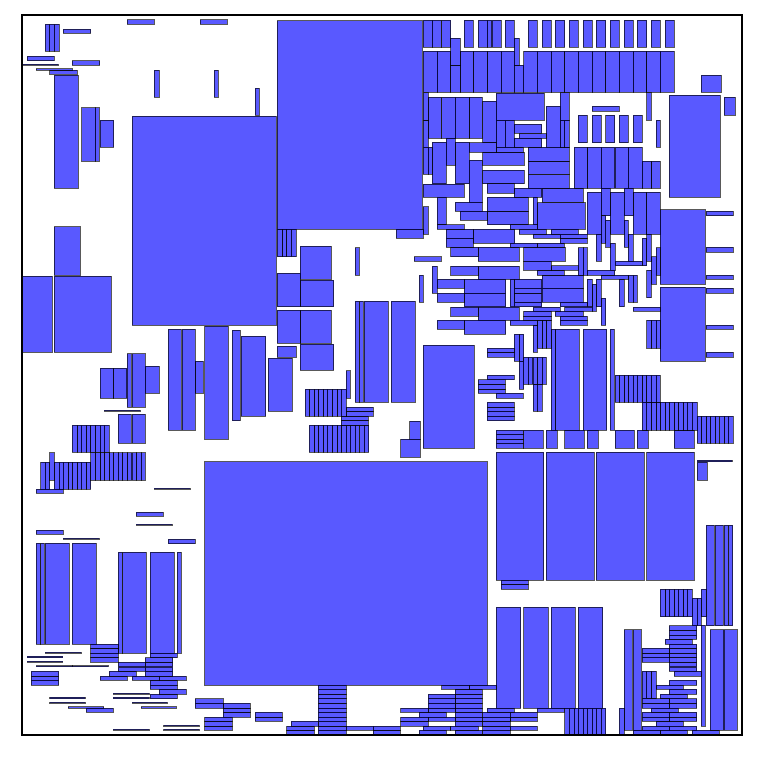} }
		\end{minipage}
	\end{minipage}
	\caption{Macro placement visualization for $adaptec2$ benchmark comparing RollPlace with five recent methods. Blue rectangles denote macros. The placement quality is evaluated through HPWL value, Lower HPWL values indicate superior performance.
	}
	\label{Fig:Compadaptec2}
\end{figure*}

We also compare the visualizations of optimal layouts for circuit $adaptec2$ between RollPlace and other methods, as shown in Fig.~\ref{Fig:Compadaptec2}.

Additionally, we report the time consumption of each algorithm in the macro placement task. The time consumption by ChiPFormer only includes the fine-tuning time and excludes the training time on offline placement data. For MaskPlace, the reported time only includes the time required to deploy 128 macros under default conditions. DREAMPlace solves the Poisson equation and uses the deep learning toolkit to accelerate the solution process, so its runtime is typically measured in seconds. As shown in Table~\ref{tab:time_consumption_minutes}, the runtime of our method remains within an acceptable range compared to other AI-based algorithms. All experiments were conducted on a single machine with NVIDIA GeForce RTX 2080 Ti GPUs and an Intel(R) Xeon(R) Gold 6226R CPU @ 2.90 GHz.


\begin{table*}[htbp]
	\renewcommand{\arraystretch}{1.2}
	\centering
	\caption{Comparison of time consumption (in minutes) for macro placement across all algorithms. DREAMPlace leverages GPU acceleration to solve the placement problem. For time consumption metrics, a smaller value indicates superior performance.}
	\label{tab:time_consumption_minutes}
	\resizebox{ 0.8\textwidth}{!}{
		\begin{tabular}{c c c c c c c c c c} 
			\toprule
			\multicolumn{2}{c}{Method} & adaptec1 & adaptec2 & adaptec3 & adaptec4 & bigblue1 & bigblue2 & bigblue3 & bigblue4  \\
			\midrule
			\multirow{1}{*}{Traditional} & DREAMPlace & $0.34$ & $0.45$ & $0.61$ & $0.70$ & $0.34$ & $6.76$ & $1.57$ & $2.49$ \\
			\hline
			\multirow{6}{*}{AI-based} & MaskPlace & $120$ & $172$ & $132$ & $119$ & $104$ & $103$ & $163$ & $132$ \\
			& ChiPFormer & $16$ & $21$ & $22$ & $21$ & $19$ & $19$ & $22$ & $29$\\
			& WireMask-EA & $184$ & $71$ & $44$ & $78$ & $31$ & $80$ & $231$ & $324$  \\
			& EfficientPlace & $71$ & $116$ & $132$ & $236$ & $83$ & $30$ & $255$ & $123$  \\
			& RollPlace & $35$ & $61$ & $57$ & $80$ &  $30$ & $38$ & $155$ & $156$ \\
			& RollPlace(1k) & $71$ & $136$ & $102$ & $166$ & $68$ & $58$ & $237$ & $236$ \\
			
			\bottomrule
		\end{tabular}
	}
\end{table*}


\subsection{Congestion Analysis} 

In addition to wirelength optimization, congestion and density serve as critical metrics for assessing placement quality. 
While density metrics are not explicitly reported here due to RollPlace's strict non-overlap hard constraints during placement, we provide a comprehensive congestion analysis. 
Table~\ref{TableCongesion} compares placement congestion metrics across state-of-the-art methods, including DeepPlace~\cite{deepplace}, DREAMPlace~\cite{nonlinear9dreamplace}, MaskPlace~\cite{maskplace}, ChiPFormer~\cite{chipformer}, WireMask-EA~\cite{wiremaskbbo}, EfficientPlace~\cite{efficientplace}, and RollPlace. 
All congestion values are calculated using the RUDY algorithm~\cite{RUDY}. 

The experimental results demonstrate RollPlace's superior congestion optimization capability. 
Specifically, RollPlace achieves the best congestion values in seven out of eight benchmark circuits and consistently maintains the second-best performance in seven out of eight benchmarks. 
This performance highlights RollPlace's ability to simultaneously optimize both wirelength and congestion metrics, despite its primary focus on HPWL minimization. 
The observed congestion reduction aligns with previous findings of Shi et al.~\cite{wiremaskbbo}, confirming that effective wirelength optimization can inherently lead to improved congestion characteristics.


\begin{table*}[htbp]
	\renewcommand{\arraystretch}{1.2}
	\centering
	\caption{Comparisons of congestion metrics. We standardize the best congestion values to $1$ for a direct comparison. These metrics are calculated directly from the results in Table~\ref{TableMainResult} using the RUDY algorithm without modifications. Note that DeepPlace results are excluded due to unavailable placement data. Lower values indicate superior congestion performance. The best-performing results are highlighted in \textbf{bold}, while the second-best results are marked in \textcolor{brown}{brown}.}
	\label{TableCongesion}
	\resizebox{\textwidth}{!}{
		\begin{tabular}{c c c c c c c c c} 
			\toprule
			Method & adaptec1 & adaptec2 & adaptec3 & adaptec4 & bigblue1 & bigblue2 & bigblue3 & bigblue4  \\
			\midrule
			DREAMPlace & $3.78\pm0.5$ & $\textcolor{brown}{1.14\pm0.08}$ & $1.15\pm0.02$ & $2.06\pm0.42$ & $9.74\pm2.01$ & $3.64\pm0.07$ & $1.71\pm0.08$ & $1.83\pm0.05$ \\
			MaskPlace & $2.81\pm0.07$ & $1.71\pm0.12$ & $1.40\pm0.07$ & $1.12\pm0.04$ & $4.42\pm0.60$ & $2.24\pm0.14$ & $1.22\pm0.05$ & $1.65\pm0.11$ \\
			ChiPFormer & $4.91\pm0.61$ & $1.98\pm0.12$ & $1.14\pm0.02$ & $1.01\pm0.01$ & $1.81\pm0.03$ & $1.39\pm0.12$ & $1.30\pm0.06$ & $1.26\pm0.09$ \\
			WireMask-EA & $1.07\pm0.03$ & $1.94\pm0.15$ & $1.11\pm0.04$ & $1.06\pm0.04$ & $1.42\pm0.08$ & $1.47\pm0.06$ & $1.25\pm0.03$ & $1.21\pm0.05$ \\
			EfficientPlace & $1.06\pm0.04$ & $1.27\pm0.11$ & $1.08\pm0.03$ & $1.02\pm0.02$ & $1.52\pm0.07$ & $\boldsymbol{1.00\pm0.04}$ & $1.56\pm0.32$ & $1.11\pm0.02$ \\
			RollPlace & $\boldsymbol{1.00\pm0.04}$ & $1.14\pm0.22$ & $\textcolor{brown}{1.04\pm0.06}$ & $\boldsymbol{1.00\pm0.03}$ & $\textcolor{brown}{1.02\pm0.13}$ & $1.15\pm0.04$ & $\textcolor{brown}{1.10\pm0.09}$ & $\textcolor{brown}{1.09\pm0.06}$ \\
			RollPlace(1k) & $\textcolor{brown}{1.03\pm0.11}$ & $\boldsymbol{1.00\pm0.08}$ & $\boldsymbol{1.00\pm0.06}$ & $\textcolor{brown}{1.00\pm0.05}$ & $\boldsymbol{1.0\pm0.01}$ & $\textcolor{brown}{1.14\pm0.06}$ & $\boldsymbol{1.00\pm0.11}$ & $\boldsymbol{1.00\pm0.06}$ \\
			\bottomrule
		\end{tabular}
	}
\end{table*}

\subsection{Full Placement Analysis} 
Existing studies commonly employ DREAMPlace~\cite{nonlinear9dreamplace} to replace macros within the full placement workflow. 
However, this approach often leads to significant deviations between the final macro positions and algorithm's macro placement results. 
While DREAMPlace demonstrates strength in macro placement, such workflows inadequately assess the efficacy of dedicated macro layout optimization algorithms. 
To address this, inspired by DeepPlace~\cite{deepplace} and WireMask-EA, we propose a two-stage placement strategy for full placement tasks: 
(1) exclusively optimizing macro placement without considering any standard cells, 
(2) fixing the optimized macro positions, then employing DREAMPlace to place the remaining standard cells and perform legalization. 
We exclude $bigblue2$ and $bigblue4$ from the full placement experiments because their macro placement configurations did not include all available macros.

\begin{figure}[htbp]
	
	
	\def \wTwofig {0.49\textwidth}
	
	\begin{minipage}{\columnwidth}
		\centering
		
		\begin{minipage}{\wTwofig}
			\subfloat[Macro placement]{ \includegraphics[width=\textwidth]{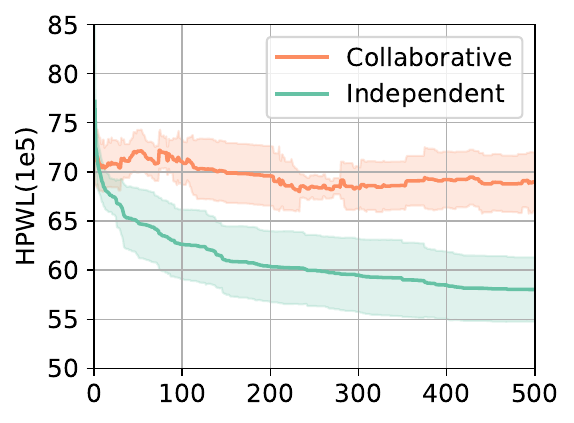}\label{Fig:WorkflowCompare-a} }
		\end{minipage}
		\hfill
		\begin{minipage}{\wTwofig}
			\subfloat[Full placement]{ \includegraphics[width=\textwidth]{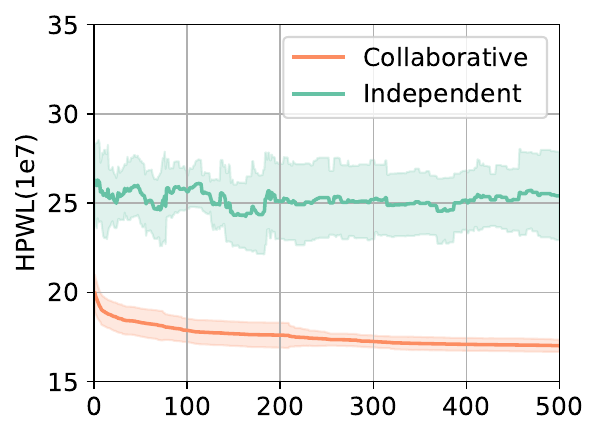}\label{Fig:WorkflowCompare-b}
			}
		\end{minipage}
	\end{minipage}
	\caption{Comparison of HPWL relationship between macro placement and full placement on $adaptec3$ under different workflows. The green curve reflects HPWL trends across running steps in the independent workflow, whereas the orange curve captures HPWL trends in the proposed collaborative optimization process.}
	\label{Fig:WorkflowCompare}
\end{figure}

Through extensive experimentation with the previous independent workflow, we find that the HPWL values of full placement and macro placement are not strictly proportional.
Specifically, a smaller macro placement HPWL value does not invariably lead to a better full placement HPWL value. 
This phenomenon is illustrated by the green curve in Fig.~\ref{Fig:WorkflowCompare}, which shows the evolving relationship between macro placement and full placement HPWL throughout the optimization process. 
This weaker correlation in the independent workflow served as the impetus for developing our novel collaborative optimization strategy.

The collaborative workflow proceeds as follows:
(1) Place macros with RollPlace while omitting macro HPWL calculations,
(2) after all macros are placed on the canvas, place all standard cells using DREAMPlace, and
(3) calculate full placement HPWL value to replace the macro HPWL value and propagate this metric to the search tree for iterative refinement.
As illustrated by the orange curve in Fig.~\ref{Fig:WorkflowCompare}, this approach establishes a positive correlation between macro HPWL and full placement HPWL. 
The results indicate that when optimizing HPWL for full placement, our collaborative method concurrently improves HPWL for macro placement, yielding more consistent and predictable optimization outcomes.


\begin{table*}[htbp]
	\renewcommand{\arraystretch}{1.2}
	\centering
	\caption{Comparisons of full placement HPWL results ($\times10^7$). Lower HPWL values indicate superior performance. Data for DeepPlace~\cite{deepplace} are obtained from~\cite{wiremaskbbo}, while RollPlace results are generated via our full placement workflow. Results for other methods are derived from the workflow described above. The best-performing results are highlighted in \textbf{bold}, while the second-best results are marked in \textcolor{brown}{brown}.}
	\label{TableFullPlacement}
	\resizebox{0.8\textwidth}{!}{
		\begin{tabular}{c c c c c c c} 
			\toprule
			Method & adaptec1 & adaptec2 & adaptec3 & adaptec4 & bigblue1 & bigblue3 \\
			\midrule
			DeepPlace & $8.28$ & $12.33$ & $23.24$ & $23.40$ & $14.10$ & $46.86$ \\
			MaskPlace & $11.37\pm0.13$ & $13.29\pm0.58$ & $26.88\pm1.50$ & $25.11\pm0.52$ & $11.30\pm0.81$ & $54.94\pm1.32$ \\
			ChiPFormer & $8.44\pm0.54$ & $12.05\pm0.93$ & $27.05\pm1.14$ & $23.76\pm0.68$ & $9.70\pm0.26$ & $52.90\pm2.50$ \\
			WireMask-EA & $10.96\pm0.30$ & $13.55\pm2.10$ & $27.02\pm1.50$ & $25.96\pm2.07$ & $12.28\pm2.67$ & $57.90\pm6.33$ \\
			EfficientPlace & $11.08\pm0.52$ & $12.98\pm0.37$ & $27.90\pm1.12$ & $26.85\pm1.73$ & $11.14\pm0.51$ & $53.63\pm11.01$ \\
			DREAMPlace & $\textcolor{brown}{7.43}$ & $\textcolor{brown}{8.42}$ & $\textcolor{brown}{20.79}$ & $\textcolor{brown}{17.90}$ & $\textcolor{brown}{9.03}$ & $\textcolor{brown}{30.71}$ \\
			RollPlace & $\boldsymbol{7.28\pm0.06}$ & $\boldsymbol{7.62\pm0.11}$ & $\boldsymbol{16.99\pm0.20}$ & $\boldsymbol{17.21\pm0.37}$ & $\boldsymbol{8.84\pm0.04}$ & $\boldsymbol{29.51\pm0.16}$ \\
			\bottomrule
		\end{tabular}
	}
\end{table*}

\subsection{Full Placement Experimental Results}

To evaluate the effectiveness of RollPlace in full placement tasks, we perform extensive experiments on the ISPD 2005 benchmarks~\cite{nam2005ispd2005}, comparing RollPlace with five recent methods, including MaskPlace~\cite{maskplace}, ChiPFormer~\cite{chipformer}, WireMask-EA~\cite{wiremaskbbo}, and EfficientPlace~\cite{efficientplace}. 
While all these methods except DREAMPlace employ diverse workflows, they exhibit a shared limitation: the absence of collaborative optimization between macro placement and standard cell placement. To ensure an equitable comparison, we unify their evaluation under a standardized framework comprising four steps:
(1) Optimize macro placement until the predefined run steps,
(2) place standard cells with DREAMPlace, leveraging the finalized macro layout,
(3) perform legalization to ensure design rule compliance, and
(4) calculate the full placement HPWL as the final metric.
The results of DREAMPlace come from placing standard cells with DREAMPlace based on the original macro positions. In contrast, RollPlace utilizes our proposed collaborative optimization workflow and initializes with the original macro positions from the ISPD 2005 benchmarks. The results of other algorithms come from the above process.

Table~\ref{TableFullPlacement} reports results for mixed-size placement, where the objective is to minimize the HPWL of all cells (macros + standard cells). The experimental results demonstrate RollPlace's superior performance in full placement improvement. 
Our method consistently achieves state-of-the-art results across all benchmark circuits, outperforming recent approaches. 
This remarkable performance not only validates the effectiveness of our collaborative optimization strategy but also highlights RollPlace's potential as a robust solution for modern placement challenges. 
The consistent improvement across diverse benchmarks suggests that the integration of macro and standard cell placement optimization in a unified framework is crucial for achieving optimal placement quality.

\subsection{End-to-End Evaluation}

The primary objective of the EDA workflow is to optimize the final PPA metrics. To evaluate the effectiveness of our method within a complete EDA workflow, we adopt open-source datasets. Specifically, we use 19 benchmarks from ChiPBench~\cite{chipbench} at the NanGate45 (NG45) technology node. Following the ChiPBench methodology, we construct an end-to-end evaluation process using OpenROAD~\cite{kahng2021openroad} and report the final metrics.

Since ML-based algorithms, such as WireMask-EA and EfficientPlace, are implemented based on the Bookshelf file format, it is necessary to convert LEF/DEF files into the Bookshelf format during the macro placement stage. Following ChiPBench, we employ the OpenROAD flow for all steps prior to macro placement and then convert the intermediate results into the Bookshelf format. After completing macro placement, the results are converted back into the LEF/DEF format, and the subsequent steps are performed using OpenROAD. In contrast, DREAMPlace and Hier-RTLMP natively support the LEF/DEF format, eliminating the need for format conversion.

The evaluation reports two categories of metrics. The final design PPA metrics include worst negative slack (WNS (ns)), total negative slack (TNS (ns)), the number of violating paths  (NVP), Power (nW), and Area ($\mu m^2$). The intermediate surrogate metrics include Congestion ($\%$), Wirelength ($\mu m$), HPWL ($\mu m$), and MacroHPWL ($\mu m$). For detailed statistics of each benchmark and the complete evaluation flow, please refer to~\cite{kahng2021openroad,chipbench}.

\begin{figure*}[hbp]
	
	
	\def \wSevenfig {0.19\textwidth}
	
	\begin{minipage}{\textwidth}
		\centering
		
		\begin{minipage}{\wSevenfig}
			\subfloat[WireMask-EA \\ Wirelength=3374605 \\ WNS=-1.459 \\ TNS=-375.086]{ \includegraphics[width=\textwidth]{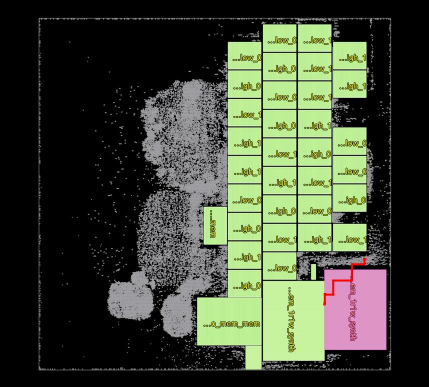} }
		\end{minipage}
		\hfill
		\begin{minipage}{\wSevenfig}
			\subfloat[EfficientPlace \\ Wirelength=3722790 \\ WNS=--1.362 \\ TNS=-695.666 ]{ \includegraphics[width=\textwidth]{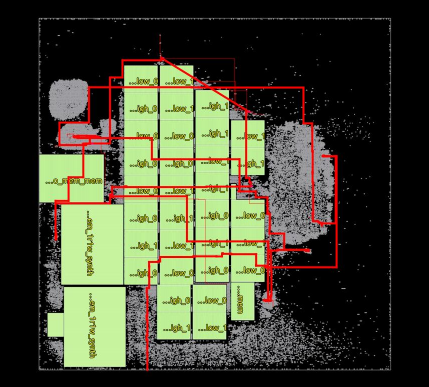} }
		\end{minipage}
		\hfill
		\begin{minipage}{\wSevenfig}
			\subfloat[DREAMPlace \\ Wirelength=3467860 \\ WNS=-1.400 \\ TNS=-455.777]{ \includegraphics[width=\textwidth]{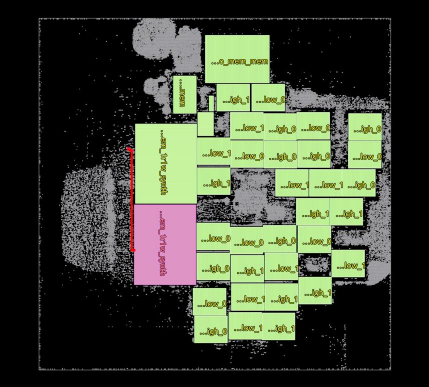} }
		\end{minipage}
		\hfill
		\begin{minipage}{\wSevenfig} 
			\subfloat[Hier-RTLMP \\ Wirelength=3737052 \\ WNS=-1.815 \\ TNS=-883.540]{ \includegraphics[width=\textwidth]{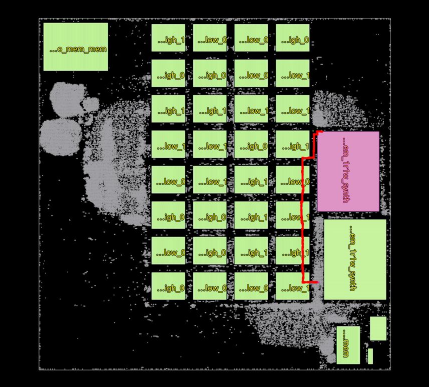} }
		\end{minipage}
		\hfill
		\begin{minipage}{\wSevenfig} 
			\subfloat[RollPlace \\ Wirelength=3138932 \\ WNS=-2.076 \\ TNS=-361.105]{ \includegraphics[width=\textwidth]{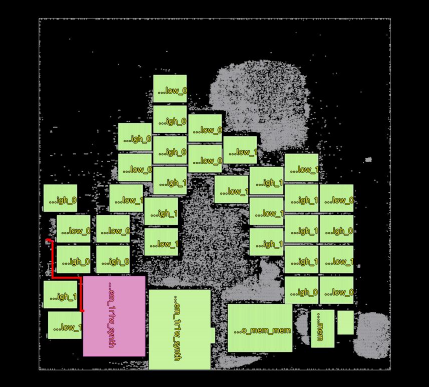} }
		\end{minipage}
	\end{minipage}
	\caption{Final placement results of different methods for the $bp\_fe38$ benchmark. For key performance metrics, a smaller WireLength indicates better routing efficiency, while larger values of WNS and TNS reflect superior timing performance.
	}
	\label{Fig:CompBpfe38}
\end{figure*}

Different from the workflow used for the ISPD2005 benchmarks, this study employs a modified procedure. First, hMetis~\cite{hMetis} is applied to cluster standard cells into 64 partitions. For each cluster, the pins of its constituent standard cells are mapped to the geometric center of the cluster’s bounding box. These clustered partitions are then treated as macros and subjected to the macro placement task. After that, the placement results are converted back to the LEF/DEF format, and the subsequent stages of the design flow are executed using OpenROAD. All experimental results are summarized in Table~\ref{tab:metrics_data}. The final placement results for the $bp\_fe38$ benchmark across all algorithms are illustrated in Fig.~\ref{Fig:CompBpfe38}.

To provide an intuitive comparison of how different algorithms affect various performance metrics, we ranked the five algorithms for each metric (with 1 representing the best and 5 the worst). The average ranking of each algorithm across all benchmarks is then computed. The final average ranking results are summarized in Table~\ref{tab:metrics_rank}.  
Two algorithms demonstrate relatively strong overall performance. As shown in Table~\ref{tab:metrics_rank}, our proposed RollPlace achieves the best average ranking in four metrics: HPWL, congestion, wirelength, and area. In contrast, Hier-RTLMP achieves the best average ranking in five other metrics: regularity, WNS, TNS, NVP, and power. This indicates that our algorithm provides significant advantages in tasks related to wirelength optimization, while Hier-RTLMP excels at capturing design dataflow. Furthermore, our method achieves competitive results in regularity, WNS, TNS, and power, suggesting that it also offers certain advantages in timing optimization under the given workflow.


\begin{table*}[htbp]
	\renewcommand{\arraystretch}{1.2}
	\centering
	\caption{Average rankings of final design PPA metrics across all benchmarks. For all metrics, a smaller ranking value indicates better performance (with 1 representing the best performance and 5 corresponding to the worst). The best-performing results are highlighted in \textbf{bold}, while the second-best results are marked in \textcolor{brown}{brown}.}
	\label{tab:metrics_rank}
	
	\adjustbox{max width=0.85\textwidth}{
		\begin{tabular}{ccccccccccc}
			\toprule
			Method & MacroHPWL & Regularity & HPWL & Congestion & Wirelength & WNS & TNS & NVP & Power & Area \\
			\midrule
			
			WireMaks-EA & $\boldsymbol{2.42}$ & $3.74$ & $3.84$ & $4.00$ & $4.11$ & $3.79$ & $3.58$ & $3.58$ & $3.63$ & $3.84$ \\
			EfficientPlace & $\textcolor{brown}{2.47}$ & $3.53$ & $3.79$ & $3.58$ & $3.58$ & $3.11$ & $3.58$ & $3.42$ & $3.26$ & $3.53$ \\
			DREAMPlace & $2.63$ & $3.47$ & $2.74$ & $\textcolor{brown}{2.74}$ & $2.74$ & $2.89$ & $2.95$ & $\textcolor{brown}{2.47}$ & $2.79$ & $3.05$ \\
			Hier-RTLMP & $3.63$ & $\boldsymbol{1.37}$ & $\textcolor{brown}{2.47}$ & $2.79$ & $\textcolor{brown}{2.58}$ & $\boldsymbol{2.37}$ & $\boldsymbol{2.05}$ & $\boldsymbol{1.79}$ & $\boldsymbol{1.74}$ & $\textcolor{brown}{2.32}$ \\
			RollPlace & $3.84$ & $\textcolor{brown}{2.89}$ & $\boldsymbol{2.16}$ & $\boldsymbol{1.79}$ & $\boldsymbol{2.00}$ & $\textcolor{brown}{2.84}$ & $\textcolor{brown}{2.84}$ & $2.79$ & $\textcolor{brown}{2.47}$ & $\boldsymbol{2.26}$ \\
			\bottomrule
			
		\end{tabular}
	}
\end{table*}

\begin{table*}[htbp] 
	\centering
	\caption{Comparisons of the final design PPA metrics and intermediate surrogate metrics. All designs are synthesized and implemented using the NanGate45 open-source technology library. The best-performing results are highlighted in \textbf{bold}, while the second-best results are marked in \textcolor{brown}{brown}.}
	\label{tab:metrics_data}
	
	\adjustbox{max width=0.85\textwidth}{
		\begin{tabular}{cccccccccccc}
			\hline
			\multirow{3}{*}{Benchmark} & \multirow{3}{*}{Method} & \multicolumn{5}{c}{Intermediate Metrics} & \multicolumn{5}{c}{PPA Metrics} \\
			\cmidrule(lr){3-7} \cmidrule(lr){8-12}
			&  & \multicolumn{3}{c}{Placement Metrics} & \multicolumn{2}{c}{Route Metrics} & \multicolumn{3}{c}{Timing Performance}  & \multirow{2}{*}{Power $\downarrow$} & \multirow{2}{*}{Area $\downarrow$} \\
			\cmidrule(lr){3-5} \cmidrule(lr){6-7} \cmidrule(lr){8-10}
			&  & MacroHPWL $\downarrow$ & Regularity $\downarrow$ & HPWL $\downarrow$ & Congestion $\downarrow$ & Wirelength $\downarrow$ & WNS $\uparrow$ & TNS $\uparrow$ & NVP $\downarrow$  &  &  \\
			\hline
			\multirow{5}{*}{ariane133} & WireMask-EA & $\textcolor{brown}{155475}$ & $140896$ & $6507282$ & $0.258$ & $7667688$ & $-0.298$ & $-416.014$ & $2404$ & $0.352$ & $385588$ \\
			& EfficientPlace & $\boldsymbol{149649}$ & $92658$ & $6391101$ & $0.255$ & $7585750$ & $\boldsymbol{-0.078}$ & $\boldsymbol{-25.788}$ & $\boldsymbol{959}$ & $0.351$ & $\textcolor{brown}{383901}$ \\
			& DREAMPlace & $190694$ & $\textcolor{brown}{87486}$ & $\boldsymbol{5806250}$ & $\boldsymbol{0.233}$ & $\boldsymbol{6949059}$ & $-0.195$ & $-113.649$ & $1656$ & $\textcolor{brown}{0.345}$ & $\boldsymbol{382488}$ \\
			& Hier-RTLMP & $568103$ & $\boldsymbol{65804}$ & $6546789$ & $0.266$ & $7693658$ & $\textcolor{brown}{-0.114}$ & $\textcolor{brown}{-48.149}$ & $\textcolor{brown}{1052}$ & $\boldsymbol{0.342}$ & $389598$ \\
			& RollPlace & $270120$ & $118370$ & $\textcolor{brown}{6334449}$ & $\textcolor{brown}{0.252}$ & $\textcolor{brown}{7495264}$ & $-0.185$ & $-144.404$ & $1673$ & $0.352$ & $384551$ \\
			\hline
			\multirow{5}{*}{ariane136}
			& WireMask-EA & $\textcolor{brown}{157147}$ & $110008$ & $6963358$ & $0.344$ & $10142768$ & $-1.812$ & $-2262.28$ & $4928$ & $0.409$ & $396286$ \\
			& EfficientPlace & $\boldsymbol{155278}$ & $128512$ & $6537015$ & $0.262$ & $7763414$ & $\boldsymbol{-0.163}$ & $\textcolor{brown}{-117.868}$ & $\textcolor{brown}{1463}$ & $\textcolor{brown}{0.378}$ & $392452$ \\
			& DREAMPlace & $195351$ & $\textcolor{brown}{84820}$ & $\boldsymbol{5735794}$ & $\boldsymbol{0.234}$ & $\boldsymbol{6942093}$ & $\textcolor{brown}{-0.170}$ & $\boldsymbol{-87.943}$ & $\boldsymbol{1203}$	 & $\boldsymbol{0.376}$ & $\boldsymbol{389967}$ \\
			& Hier-RTLMP & $525799$ & $\boldsymbol{74380}$ & $6395100$ & $0.263$ & $7588397$ & $-0.245$ & $-210.549$ & $1943$ & $0.379$ & $397250$ \\
			& RollPlace & $261447$ & $94960$ & $\textcolor{brown}{6115838}$ & $\textcolor{brown}{0.249}$ & $\textcolor{brown}{7389673}$ & $-0.392$ & $-418.008$ & $2194$ & $0.383$ & $\textcolor{brown}{391100}$ \\
			\hline
			\multirow{5}{*}{black\_parrot} 
			& WireMask-EA & $\textcolor{brown}{20396}$ & $22527$ & $9165404$ & $0.472$ & $10964822$ & $\boldsymbol{-4.554}$ & $\textcolor{brown}{-171.518}$ & $1955$ & $0.498$ & $535727$ \\
			& EfficientPlace & $\boldsymbol{19818}$ & $19594$ & $9824288$ & $0.507$ & $11755179$ & $-4.718$ & $-292.413$ & $1678$ & $0.512$ & $538959$ \\
			& DREAMPlace & $31658$ & $18167$ & $9019694$ & $0.485$ & $11255216$ & $\textcolor{brown}{-4.585}$ & $-222.481$ & $\textcolor{brown}{1545}$ & $\boldsymbol{0.485}$ & $\textcolor{brown}{530353}$ \\
			& Hier-RTLMP & $22621$ & $\boldsymbol{14169}$ & $\textcolor{brown}{8023406}$ & $\textcolor{brown}{0.420}$ & $\boldsymbol{9512238}$ & $-4.906$ & $\boldsymbol{-99.094}$ & $\boldsymbol{1390}$ & $\textcolor{brown}{0.490}$ & $\boldsymbol{530012}$ \\
			& RollPlace & $43682$ & $\textcolor{brown}{17539}$ & $\boldsymbol{8007123}$ & $\boldsymbol{0.412}$ & $\textcolor{brown}{9591922}$ & $-4.778$ & $-221.091$ & $2903$ & $\textcolor{brown}{0.490}$ & $531889$ \\
			\hline
			\multirow{5}{*}{bp\_be} 
			& WireMask-EA & $\boldsymbol{11250}$ & $4545$ & $3420429$ & $0.650$ & $4449951$ & $-0.818$ & $-72.564$ & $\textcolor{brown}{127}$ & $0.153$ & $124271$ \\
			& EfficientPlace & $\textcolor{brown}{11654}$ & $5150$ & $3309221$ & $0.6139$ & $4186004$ & $-0.799$ & $-65.932$ & $\boldsymbol{111}$ & $0.149$ & $123190$ \\
			& DREAMPlace & $16977$ & $\textcolor{brown}{4518}$ & $3377118$ & $\textcolor{brown}{0.588}$ & $4218769$ & $-0.657$ & $-51.154$ & $\boldsymbol{111}$ & $\boldsymbol{0.146}$ & $123563$ \\
			& Hier-RTLMP & $13576$ & $\boldsymbol{4333}$ & $\textcolor{brown}{3212067}$ & $0.599$ & $\textcolor{brown}{4087593}$ & $\textcolor{brown}{-0.621}$ & $\textcolor{brown}{-40.274}$ & $\boldsymbol{111}$ & $\textcolor{brown}{0.147}$ & $\textcolor{brown}{122479}$ \\
			& RollPlace & $14262$ & $5465$ & $\boldsymbol{3195611}$ & $\boldsymbol{0.568}$ & $\boldsymbol{3998973}$ & $\boldsymbol{-0.553}$ & $\boldsymbol{-39.983}$ & $\boldsymbol{111}$ & $\textcolor{brown}{0.147}$ & $\boldsymbol{122270}$ \\
			\hline
			\multirow{5}{*}{bp\_fe} 
			& WireMask-EA & $44215$ & $5198$ & $2469636$ & $0.815$ & $4230713$ & $-2.469$ & $-827.503$ & $1193$ & $0.190$ & $72056$ \\
			& EfficientPlace & $\textcolor{brown}{43871}$ & $5349$ & $2606989$ & $0.677$ & $3601380$ & $-0.969$ & $-94.775$ & $497$ & $0.182$ & $73904$ \\
			& DREAMPlace & $\boldsymbol{43729}$ & $\textcolor{brown}{4442}$ & $2277162$ & $0.615$ & $3323972$ & $-1.137$ & $-81.975$ & $393$ & $\textcolor{brown}{0.170}$ & $\textcolor{brown}{71608}$ \\
			& Hier-RTLMP & $48682$ & $\boldsymbol{4083}$ & $\boldsymbol{2051900}$ & $\boldsymbol{0.482}$ & $\boldsymbol{2631261}$ & $\boldsymbol{-0.237}$ & $\boldsymbol{-5.950}$ & $\boldsymbol{96}$ & $\boldsymbol{0.167}$ & $\boldsymbol{70165}$ \\
			& RollPlace & $51798$ & $4570$ & $\textcolor{brown}{2190620}$ & $\textcolor{brown}{0.562}$ & $\textcolor{brown}{3040182}$ & $\textcolor{brown}{-0.867}$ & $\textcolor{brown}{-77.150}$ & $\textcolor{brown}{377}$ & $0.171$ & $71837$ \\
			\hline
			\multirow{5}{*}{dft68} 
			& WireMask-EA & $\boldsymbol{179789}$ & $47109$ & $1213938$ & $0.097$ & $1354487$ & $-0.340$ & $-65.007$ & $\textcolor{brown}{277}$ & $0.232$ & $86071$ \\
			& EfficientPlace & $\textcolor{brown}{214872}$ & $46727$ & $1299050$ & $0.103$ & $1443420$ & $-0.389$ & $-68.607$ & $278$ & $0.238$ & $87152$ \\
			& DREAMPlace & $216173$ & $46434$ & $\boldsymbol{1124378}$ & $\textcolor{brown}{0.094}$ & $\boldsymbol{1266514}$ & $\textcolor{brown}{-0.315}$ & $\textcolor{brown}{-59.877}$ & $\boldsymbol{276}$ & $\boldsymbol{0.226}$ & $\textcolor{brown}{86044}$ \\
			& Hier-RTLMP & $513981$ & $\boldsymbol{34462}$ & $1752129$ & $0.135$ & $1888029$ & $\boldsymbol{-0.286}$ & $\boldsymbol{-56.865}$ & $\boldsymbol{276}$ & $0.232$ & $88144$ \\
			& RollPlace & $218560$ & $\textcolor{brown}{42853}$ & $\textcolor{brown}{1154247}$ & $\boldsymbol{0.093}$ & $\textcolor{brown}{1299421}$ & $-0.320$ & $-61.098$ & $\boldsymbol{276}$ & $\textcolor{brown}{0.231}$ & $\boldsymbol{85669}$ \\
			\hline
			\multirow{5}{*}{or1200} 
			& WireMask-EA & $103246$ & $18102$ & $884872$ & $0.073$ & $1051538$ & $-7.708$ & $-309.542$ & $149$ & $\textcolor{brown}{0.028}$ & $68982$ \\
			& EfficientPlace & $109239$ & $\textcolor{brown}{17853}$ & $888252$ & $0.072$ & $1040927$ & $-7.580$ & $-1780.920$ & $1674$ & $\textcolor{brown}{0.028}$ & $67787$ \\
			
			& DREAMPlace & $\boldsymbol{72118}$ & $26254$ & $\boldsymbol{838399}$ & $\textcolor{brown}{0.070}$ & $\textcolor{brown}{1014607}$ & $\textcolor{brown}{-6.040}$ & $-255.116$ & $\boldsymbol{115}$ & $0.042$ & $69268$ \\
			& Hier-RTLMP & $115057$ & $\boldsymbol{16540}$ & $884510$ & $0.071$ & $1027487$ & $\boldsymbol{-6.010}$ & $\boldsymbol{-215.324}$ & $\textcolor{brown}{145}$ & $\boldsymbol{0.027}$ & $\boldsymbol{66906}$ \\
			& RollPlace & $\textcolor{brown}{89244}$ & $18019$ & $\textcolor{brown}{843026}$ & $\boldsymbol{0.069}$ & $\boldsymbol{995492}$ & $-6.487$ & $\textcolor{brown}{-243.606}$ & $155$ & $\boldsymbol{0.027}$ & $\textcolor{brown}{67246}$ \\
			\hline
			\multirow{5}{*}{swerv\_wrapper} 
			& WireMask-EA & $\textcolor{brown}{89410}$ & $17455$ & $4681869$ & $0.396$ & $5559256$ & $-1.412$ & $-1344.120$ & $1729$ & $0.270$ & $232705$ \\
			& EfficientPlace & $\boldsymbol{78835}$ & $16384$ & $\textcolor{brown}{4246494}$ & $\textcolor{brown}{0.362}$ & $\textcolor{brown}{5064971}$ & $\textcolor{brown}{-0.610}$ & $\boldsymbol{-541.569}$ & $\textcolor{brown}{1432}$ & $\textcolor{brown}{0.265}$ & $\textcolor{brown}{229790}$ \\
			& DREAMPlace & $101776$ & $\textcolor{brown}{13783}$ & $4506378$ & $0.369$ & $5359494$ & $-0.844$ & $-645.860$ & $1526$ & $0.267$ & $230232$ \\
			& Hier-RTLMP & $118198$ & $16732$ & $\boldsymbol{3829279}$ & $\boldsymbol{0.330}$ & $\boldsymbol{4603628}$ & $\boldsymbol{-0.599}$ & $\textcolor{brown}{-583.211}$ & $1436$ & $\boldsymbol{0.262}$ & $\boldsymbol{226460}$ \\
			& RollPlace & $109254$ & $\boldsymbol{13554}$ & $4651057$ & $0.396$ & $5543365$ & $-0.987$ & $-737.212$ & $\boldsymbol{1409}$ & $0.283$ & $231627$ \\
			\hline
			\multirow{5}{*}{bp\_multi} 
			& WireMask-EA & $\textcolor{brown}{24117}$ & $18741$ & $4829535$ & $\textcolor{brown}{0.379}$ & $\textcolor{brown}{5581405}$ & $\boldsymbol{-5.483}$ & $\boldsymbol{-2628.98}$ & $\boldsymbol{9225}$ & $\boldsymbol{0.535}$ & $\textcolor{brown}{263234}$\\
			& EfficientPlace & $24897$ & $\textcolor{brown}{14363}$ & $\textcolor{brown}{5244707}$ & $0.407$ & $5946394$ & $-5.758$ & $-3907.67$ & $10065$ & $\textcolor{brown}{0.538}$ & $266985$ \\
			& DREAMPlace & $34078$ & $16973$ & $5312844$ & $0.418$ & $6146687$ & $\textcolor{brown}{-5.591}$ &  $-3226.07$ & $9448$ & $0.539$ & $266624$ \\
			& Hier-RTLMP & $\boldsymbol{22784}$ & $\boldsymbol{11794}$ & $\boldsymbol{4765293}$ & $\boldsymbol{0.376}$ & $\boldsymbol{5485683}$ & $-5.785$ &  $\textcolor{brown}{-2740.43}$ & $\textcolor{brown}{9409}$ & $\boldsymbol{0.535}$ & $\boldsymbol{262806}$ \\
			& RollPlace & $46036$ & $16243$ & $5456017$ & $0.408$ & $6178172$ & $-5.647$ &  $-2946.30$ & $9695$ & $\textcolor{brown}{0.538}$ & $266525$ \\
			\hline
			\multirow{5}{*}{vga\_lcd} 
			& WireMask-EA & $\textcolor{brown}{53572}$ & $\textcolor{brown}{32980}$ & $\boldsymbol{1461215}$ & $0.154$ & $3036396$ & $-5.387$ & $\textcolor{brown}{-1839.45}$ & $\textcolor{brown}{3774}$ & $\boldsymbol{0.180}$ & $262468$ \\
			& EfficientPlace & $\boldsymbol{45907}$ & $39213$ & $\textcolor{brown}{1463868}$ & $0.146$ & $2975637$ & $-4.328$ & $-4834.09$ & $9307$ & $\textcolor{brown}{0.184}$ & $271819$ \\
			& DREAMPlace & $57832$ & $49551$ & $1663069$ & $\boldsymbol{0.123}$ & $\boldsymbol{2511081}$ & $-1.650$ & $-6766.29$ & $21090$ & $0.194$ & $\textcolor{brown}{255125}$ \\
			& Hier-RTLMP & $101530$ & $\boldsymbol{28761}$ & $1717906$ & $\textcolor{brown}{0.132}$ & $\textcolor{brown}{2684816}$ & $\boldsymbol{-0.982}$ & $\boldsymbol{-431.846}$ & $\boldsymbol{3236}$ & $0.191$ & $261276$ \\
			& RollPlace & $72707$ & $49513$ & $1836099$ & $0.137$ & $2782021$ & $\textcolor{brown}{-1.060}$ & $-6359.71$ & $20982$ & $0.193$ & $\boldsymbol{254758}$ \\
			\hline
			\multirow{5}{*}{mor1kx} 
			& WireMask-EA & $78270$ & $20805$ & $1983374$ & $0.408$ & $2588734$ & $-1.384$ & $-11159.8$ & $10198$ & $0.254$ & $155424$ \\
			& EfficientPlace & $77209$ & $20992$ & $1964136$ & $0.405$ & $2564338$ & $-1.346$ & $\textcolor{brown}{-10474.3}$ & $\textcolor{brown}{10092}$ & $0.257$ & $\textcolor{brown}{155086}$ \\
			& DREAMPlace & $\boldsymbol{37540}$ & $35424$ & $\boldsymbol{1820175}$ & $\boldsymbol{0.382}$ & $\boldsymbol{2438671}$ & $-1.368$ & $-11109.4$ & $10134$ & $0.257$ & $155713$ \\
			& Hier-RTLMP & $73492$ & $\boldsymbol{19766}$ & $1950767$ & $0.407$ & $2581561$ & $\textcolor{brown}{-1.341}$ & $\boldsymbol{-10384.2}$ & $\boldsymbol{10068}$ & $\textcolor{brown}{0.253}$ & $\boldsymbol{154708}$ \\
			& RollPlace & $\textcolor{brown}{72364}$ & $\textcolor{brown}{20431}$ & $\textcolor{brown}{1928368}$ & $\textcolor{brown}{0.399}$ & $\textcolor{brown}{2546238}$ & $\boldsymbol{-1.340}$ & $-10817.2$ & $10175$ & $\boldsymbol{0.252}$ & $155254$ \\
			\hline
			\multirow{5}{*}{ethernet} 
			& WireMask-EA & $31428$ & $22666$ & $871709$ & $0.288$ & $1100657$ & $-0.192$ & $-5.647$ & $90$ & $\boldsymbol{0.117}$ & $96216$ \\
			& EfficientPlace & $32848$ & $18821$ & $875794$ & $0.288$ & $1102948$ & $-0.245$ & $-9.088$ & $109$ & $\boldsymbol{0.117}$ & $\boldsymbol{96089}$ \\
			& DREAMPlace & $\boldsymbol{27200}$ & $19870$ & $886057$ & $0.294$ & $1122677$ & $\boldsymbol{-0.094}$ & $\textcolor{brown}{-4.248}$ & $109$ & $\textcolor{brown}{0.118}$ & $96387$ \\
			& Hier-RTLMP & $\textcolor{brown}{27959}$ & $\boldsymbol{13597}$ & $\textcolor{brown}{859826}$ & $\textcolor{brown}{0.286}$ & $\textcolor{brown}{1090616}$ & $\textcolor{brown}{-0.101}$ & $\boldsymbol{-1.138}$ & $\boldsymbol{53}$ & $\boldsymbol{0.117}$ & $96261$ \\
			& RollPlace & $35668$ & $\textcolor{brown}{15199}$ & $\boldsymbol{859207}$ & $\boldsymbol{0.285}$ & $\boldsymbol{1087237}$ & $-0.191$ & $-5.359$ & $\textcolor{brown}{88}$ & $\boldsymbol{0.117}$ & $\textcolor{brown}{96210}$ \\
			\hline
			\multirow{5}{*}{isa\_npu} 
			& WireMask-EA & $68074$ & $12058$ & $6355434$ & $0.197$ & $7844377$ & $-0.928$ & $-26.758$ & $52$ & $0.944$ & $\textcolor{brown}{998762}$ \\
			& EfficientPlace & $69582$ & $12750$ & $6321304$ & $0.199$ & $7830484$ & $\boldsymbol{-0.877}$ & $-27.015$ & $56$ & $\boldsymbol{0.941}$ & $999421$ \\
			& DREAMPlace & $122149$ & $\textcolor{brown}{11866}$ & $6460849$ & $0.203$ & $7960518$ & $-1.015$ & $\boldsymbol{-25.337}$ & $\boldsymbol{43}$ & $\boldsymbol{0.941}$ & $1000200$ \\
			& Hier-RTLMP & $\boldsymbol{39446}$ & $\boldsymbol{10354}$ & $\textcolor{brown}{6190877}$ & $\textcolor{brown}{0.196}$ & $\textcolor{brown}{7680181}$ & $-0.955$ & $-27.691$ & $52$ & $\textcolor{brown}{0.942}$ & $\boldsymbol{998084}$ \\
			& RollPlace & $\textcolor{brown}{63470}$ & $13164$ & $\boldsymbol{6166348}$ & $\boldsymbol{0.192}$ & $\boldsymbol{7641026}$ & $\textcolor{brown}{-0.912}$ & $\textcolor{brown}{-25.741}$ & $\textcolor{brown}{47}$ & $0.945$ & $999129$ \\
			\hline
			\multirow{5}{*}{ariane81} 
			& WireMask-EA & $\textcolor{brown}{149427}$ & $76314$ & $5851983$ & $0.322$ & $7173067$ & $-0.342$ & $-494.304$ & $2544$ & $0.190$ & $341238$ \\
			& EfficientPlace & $\boldsymbol{134531}$ & $73972$ & $6109744$ & $0.369$ & $8351496$ & $-0.983$ & $-2382.980$ & $3420$ & $0.194$ & $339517$ \\
			& DREAMPlace & $206821$ & $59932$ & $\boldsymbol{4848578}$ & $\boldsymbol{0.267}$ & $\boldsymbol{5949468}$ & $\boldsymbol{-0.133}$ & $\boldsymbol{-64.289}$ & $\boldsymbol{1103}$ & $0.189$ & $\textcolor{brown}{336665}$ \\
			& Hier-RTLMP & $330885$ & $\boldsymbol{53896}$ & $5440447$ & $0.304$ & $6557590$ & $\textcolor{brown}{-0.146}$ & $\textcolor{brown}{-127.447}$ & $\textcolor{brown}{1639}$ & $\textcolor{brown}{0.187}$ & $339833$ \\
			& RollPlace & $219111$ & $\textcolor{brown}{57129}$ & $\textcolor{brown}{4910179}$ & $\textcolor{brown}{0.268}$ & $\textcolor{brown}{5977030}$ & $-0.279$ & $-148.917$ & $1685$ & $\boldsymbol{0.186}$ & $\boldsymbol{336084}$ \\
			\hline
			\multirow{5}{*}{bp\_fe38} 
			& WireMask-EA & $922755$ & $19689$ & $\textcolor{brown}{3073331}$ & $\textcolor{brown}{0.404}$ & $\textcolor{brown}{3374605}$ & $-1.459$ & $\textcolor{brown}{-375.086}$ & $\boldsymbol{616}$ & $\textcolor{brown}{0.112}$ & $60643$ \\
			& EfficientPlace & $\textcolor{brown}{910340}$ & $20493$ & $3331842$ & $0.444$ & $3722790$ & $\boldsymbol{-1.362}$ & $-695.666$ & $1131$ & $\textcolor{brown}{0.112}$ & $63226$ \\
			& DREAMPlace & $\boldsymbol{836309}$ & $21139$ & $3127117$ & $0.415$ & $3467860$ & $\textcolor{brown}{-1.400}$ & $-455.777$ & $\textcolor{brown}{642}$ & $\boldsymbol{0.111}$ & $\boldsymbol{59742}$ \\
			& Hier-RTLMP & $960485$ & $\textcolor{brown}{19504}$ & $3295364$ & $0.463$ & $3737052$ & $-1.815$ & $-883.540$ & $1504$ & $\boldsymbol{0.111}$ & $62807$ \\
			& RollPlace & $931533$ & $\boldsymbol{18298}$ & $\boldsymbol{2837217}$ & $\boldsymbol{0.377}$ & $\boldsymbol{3138932}$ & $-2.076$ & $\boldsymbol{-361.105}$ & $\textcolor{brown}{642}$ & $\boldsymbol{0.111}$ & $\textcolor{brown}{59958}$ \\
			\hline
			\multirow{5}{*}{bp\_be12} 
			& WireMask-EA & $785283$ & $\boldsymbol{4018}$ & $3696067$ & $0.527$ & $4214949$ & $\textcolor{brown}{-0.638}$ & $\textcolor{brown}{-56.489}$ & $\boldsymbol{114}$ & $0.077$ & $93600$ \\
			& EfficientPlace & $799616$ & $5310$& $3672431$ & $0.525$ & $4175524$ & $-0.804$ & $-68.646$ & $\textcolor{brown}{115}$ & $0.076$ & $93727$ \\
			& DREAMPlace & $\boldsymbol{728422}$ & $6386$ & $\boldsymbol{3512250}$ & $\boldsymbol{0.486}$ & $\boldsymbol{3986750}$ & $-0.724$ & $-67.658$ & $\boldsymbol{114}$ & $\textcolor{brown}{0.075}$ & $\textcolor{brown}{91549}$ \\
			& Hier-RTLMP & $\textcolor{brown}{759742}$ & $4758$ & $\textcolor{brown}{3527781}$ & $0.532$ & $4281669$ & $-0.811$ & $-73.532$ & $\boldsymbol{114}$ & $\boldsymbol{0.074}$ & $\boldsymbol{88021}$ \\
			& RollPlace & $801407$ & $\textcolor{brown}{4353}$ & $3539799$ & $\textcolor{brown}{0.506}$ & $\textcolor{brown}{4005604}$ & $\boldsymbol{-0.568}$ & $\boldsymbol{-51.651}$ & $\boldsymbol{114}$ & $0.076$ & $93041$ \\
			\hline
			\multirow{5}{*}{bp68} 
			& WireMask-EA & $\boldsymbol{1517203}$ & $\textcolor{brown}{65369}$ & $19213836$ & $0.696$ & $22421712$ & $\textcolor{brown}{-2.253}$ & $\textcolor{brown}{-1102.22}$ & $951$ & $0.155$ & $305872$ \\
			& EfficientPlace & $1617941$ & $68809$ & $19400373$ & $0.691$ & $22280943$ & $-3.717$ & $-2463.41$ & $1682$ & $0.157$ & $311245$ \\
			& DREAMPlace & $\textcolor{brown}{1519611}$ & $71432$ & $14404929$ & $0.514$ & $16163701$ & $-3.029$ & $-1161.74$ & $574$ & $0.154$ & $291142$ \\
			& Hier-RTLMP & $1636220$ & $\boldsymbol{51848}$ & $\textcolor{brown}{12356892}$ & $\textcolor{brown}{0.466}$ & $\textcolor{brown}{14360494}$ & $\boldsymbol{-2.152}$ & $\boldsymbol{-684.921}$ & $\boldsymbol{557}$ & $\boldsymbol{0.149}$ & $\textcolor{brown}{281856}$ \\
			& RollPlace & $1559154$ & $72057$ & $\boldsymbol{12007845}$ & $\boldsymbol{0.434}$ & $\boldsymbol{13937557}$ & $-3.098$ & $-1267.47$ & $\textcolor{brown}{573}$ & $\textcolor{brown}{0.152}$ & $\boldsymbol{275980}$ \\
			\hline
			\multirow{5}{*}{swerv\_wrapper43} 
			& WireMask-EA & $\boldsymbol{138792}$ & $25677$ & $4514285$ & $0.219$ & $5276355$ & $-0.613$ & $-452.786$ & $1070$ & $0.259$ & $228524$ \\
			& EfficientPlace & $\textcolor{brown}{149814}$ & $\boldsymbol{20833}$ & $4326955$ & $0.217$ & $5106941$ & $\boldsymbol{-0.482}$ & $\boldsymbol{-316.883}$ & $\boldsymbol{987}$ & $\textcolor{brown}{0.255}$ & $\textcolor{brown}{226014}$ \\
			& DREAMPlace & $330670$ & $35929$ & $4671119$ & $0.225$ & $5412770$ & $-0.841$ & $-490.785$ & $1248$ & $0.262$ & $231113$ \\
			& Hier-RTLMP & $198090$ & $\textcolor{brown}{23576}$ & $\boldsymbol{4208196}$ & $\textcolor{brown}{0.211}$ & $\boldsymbol{4958294}$ & $\textcolor{brown}{-0.534}$ & $\textcolor{brown}{-385.210}$ & $\textcolor{brown}{1057}$ & $0.256$ & $\boldsymbol{225769}$ \\
			& RollPlace & $201904$ & $30513$ & $\textcolor{brown}{4244105}$ & $\boldsymbol{0.209}$ & $\textcolor{brown}{5029937}$ & $-0.642$ & $-598.949$ & $1467$ & $\boldsymbol{0.254}$ & $227217$ \\
			\hline
			\multirow{5}{*}{bp\_multi57} 
			& WireMask-EA & $1705762$ & $39733$ & $42025975$ & $3.208$ & $47287425$ & $-6.406$ & $-3801.37$ & $1285$ & $0.184$ & $478237$ \\
			& EfficientPlace & $1041030$ & $42733$ & $6880015$ & $0.537$ & $7676157$ & $-2.551$ & $-628.698$ & $408$ & $0.107$ & $209585$ \\
			& DREAMPlace & $\boldsymbol{523787}$ & $37848$ & $7484736$ & $0.584$ & $8353784$ & $-3.228$ & $-982.560$ & $471$ & $0.107$ & $214791$ \\
			& Hier-RTLMP & $\textcolor{brown}{739345}$ & $\boldsymbol{25468}$ & $\boldsymbol{5677569}$ & $\boldsymbol{0.467}$ & $\boldsymbol{6343384}$ & $\boldsymbol{-1.899}$ & $\boldsymbol{-384.005}$ & $\boldsymbol{347}$ & $\boldsymbol{0.105}$ & $\boldsymbol{204078}$ \\
			& RollPlace & $745533$ & $\textcolor{brown}{36651}$ & $\textcolor{brown}{6125984}$ & $\textcolor{brown}{0.474}$ & $\textcolor{brown}{6739992}$ & $\textcolor{brown}{-2.101}$ & $\textcolor{brown}{-492.077}$ & $\textcolor{brown}{377}$ & $\textcolor{brown}{0.106}$ & $\textcolor{brown}{205830}$ \\
			\hline
		\end{tabular}
	}
\end{table*}

\captionsetup[subfigure]{justification=centering}


\section{Conclusion} \label{Sec:Conclusion}

In this paper, we propose RollPlace, a novel macro placement improvement framework that systematically enhances initial layouts through iterative local adjustments. 
By integrating Monte Carlo rollout search with heuristic functions, RollPlace achieves significant improvements in overall layout quality. 
It demonstrates remarkable versatility by seamlessly integrating with state-of-the-art standard cell placement tools such as DREAMPlace, thereby facilitating effective optimization of full placement scenarios. 
Extensive experiments show that RollPlace consistently outperforms recent RL and black-box optimization based algorithms, while exhibiting superior robustness across diverse benchmarks. 
Furthermore, our analysis highlights the inherent limitations of one-time generation  in current RL-based methods. RollPlace addresses these challenges through its iterative refinement strategy, offering a promising pathway for future advancements in this domain.

Despite these advancements, several limitations warrant consideration. 
First, similar to recent RL-based methods and the WireMask-EA approach, RollPlace encounters scalability challenges when processing benchmarks with an exceptionally large number of macros. 
Second, RollPlace employs only a simple wire mask as a heuristic function, without considering its impact on the quality of full placement. 
Third, within the full placement workflow, evaluation relies solely on metrics from the final placement results, and this methodology requires substantial computational resources due to the need for complete standard-cell placement.
Future research will address these limitations by developing more sophisticated heuristic functions and more efficient standard-cell placement strategies.
Finally, since RTL dataflow is not considered, RollPlace does not achieve state-of-the-art performance in terms of WNS, TNS, and some other metrics. Incorporating timing information into the framework offers a promising direction for further improvement.


\bibliographystyle{IEEEtran}
\bibliography{reference/MyRef}

\end{document}